\documentclass[runningheads]{llncs}

\usepackage{graphicx}

\usepackage{tikz}
\usepackage{comment}
\usepackage{amsmath,amssymb} % define this before the line numbering.
\usepackage{color}

\usepackage{hyperref}
\hypersetup{
    colorlinks=true,
}
\usepackage{booktabs}
\usepackage{bm}
\usepackage{xfrac}
\usepackage{xcolor}
\usepackage{wrapfig}
\usepackage[]{algorithm2e, algpseudocode}
\usepackage{listings}
\usepackage{multirow}
\usepackage{stfloats}
\usepackage[normalem]{ulem} % delete this later
\usepackage{balance}

\newcommand{\R}{\mathbb{R}}
\newcommand*{\MyToprule}{%
  \cmidrule[\heavyrulewidth]%
}
\newcommand*{\MyMidrule}{%
  \cmidrule%
}

\def\etal{\emph{et al}.~}

\usepackage[capitalize]{cleveref}
\crefname{section}{Sec.}{Secs.}
\Crefname{section}{Section}{Sections}
\Crefname{table}{Table}{Tables}
\crefname{table}{Tab.}{Tabs.}

\begin{document}
% \linenumbers
\pagestyle{headings}
\mainmatter
\def\ECCVSubNumber{7539}  % Insert your submission number here

\title{PRIME: A Few Primitives Can Boost Robustness to Common Corruptions} % Replace with your title

\begin{comment}
\titlerunning{ECCV-22 submission ID \ECCVSubNumber} 
\authorrunning{ECCV-22 submission ID \ECCVSubNumber} 
\author{Anonymous ECCV submission}
\institute{Paper ID \ECCVSubNumber}
\end{comment}
%******************

% \begin{comment}

% \titlerunning{PRIME Augmentations}

\author{Apostolos Modas\thanks{The first two authors contributed equally to this work.}\inst{1} \and
Rahul Rade\protect\footnotemark[1]\inst{2} \and
Guillermo Ortiz-Jim{\'e}nez\inst{1} \and \\
Seyed-Mohsen Moosavi-Dezfooli\inst{3} \and
Pascal Frossard\inst{1}
}
\authorrunning{A. Modas et al.}
\institute{Ecole Polytechnique F{\'e}d{\'e}rale de Lausanne (EPFL), Switzerland
\and
ETH Z{\"u}rich, Switzerland
\and
Imperial College London, United Kingdom
}
% \end{comment}

%******************
\maketitle
\setcounter{footnote}{0}

\begin{abstract}
Despite their impressive performance on image classification tasks, deep networks have a hard time generalizing to unforeseen corruptions of their data. To fix this vulnerability, prior works have built complex data augmentation strategies, combining multiple methods to enrich the training data. However, introducing intricate design choices or heuristics makes it hard to understand which elements of these methods are indeed crucial for improving robustness. In this work, we take a step back and follow a principled approach to achieve robustness to common corruptions. We propose PRIME, a general data augmentation scheme that relies on simple yet rich families of max-entropy image transformations. PRIME outperforms the prior art in terms of corruption robustness, while its simplicity and plug-and-play nature enable combination with other methods to further boost their robustness. We analyze PRIME to shed light on the importance of the mixing strategy on synthesizing corrupted images, and to reveal the robustness-accuracy trade-offs arising in the context of common corruptions. Finally, we show that the computational efficiency of our method allows it to be easily used in both on-line and off-line data augmentation schemes\footnote{Our code is available at~{\footnotesize\url{https://github.com/amodas/PRIME-augmentations}}}.
\end{abstract}

%%%%%%%%%%%%%%%%%%%%%%%%%%%%%%%%%%%%%%
\section{Introduction}
\label{sec:introduction}
%%%%%%%%%%%%%%%%%%%%%%%%%%%%%%%%%%%%%%
Deep image classifiers do not work well in the presence of various types of distribution shifts~\cite{dodgekaram2016,humans2018,taori2020}. Most notably, their performance can severely drop when the input images are affected by common corruptions that are not contained in the training data, such as digital artefacts, low contrast, or blurs~\cite{corruptions2019,cbar2021}.
In general, ``common corruptions'' is an umbrella term coined to describe the set of all possible distortions that can happen to natural images during their acquisition, storage, and processing lifetime, which can be very diverse.  Nevertheless, while the space of possible perturbations is huge, the term ``common corruptions'' is generally used to refer to image transformations that, while degrading the quality of the images, still preserve their semantic information.

Building classifiers that are robust to common corruptions is far from trivial. A naive solution is to include data with all sorts of corruptions during training, but the sheer scale of all possible types of typical perturbations that might affect an image is simply too large. Moreover, the problem is per se ill-defined since there exists no formal description of all possible common corruptions.

% To overcome this issue, the research community has recently favoured increasing the ``diversity'' of the training data via data augmentation schemes~\cite{autoaugment2019,augmix2020,deepaugment2021}. Intuitively, the hope is that showing very diverse augmentations of an image to a network would increase the chance that the latter becomes invariant to some common corruptions. Still, guaranteeing a good coverage over the whole space of such corruptions is hard. Hence, the current literature relies mostly on increasing the complexity of their training pipelines by combining independently developed data augmentation strategies, to hopefully increase diversity of augmentations and achieve state-of-the-art results on different common corruptions benchmarks~\cite{ant2020,deepaugment2021,kireev2021,chenAmplitute2021,calian2021,augmax2021}. This incremental research strategy, though, has come at a cost.

To overcome this issue, the research community has recently favoured increasing the ``diversity'' of the training data via data augmentation schemes~\cite{autoaugment2019,augmix2020,deepaugment2021}. Intuitively, the hope is that showing very diverse augmentations of an image to a network would increase the chance that the latter becomes invariant to some common corruptions. Still, covering the full space of common corruptions is hard. Hence, current literature has mostly resorted to increasing the diversity of augmentations by designing intricate data augmentation pipelines, e.g., introducing DNNs for generating varied augmentations~\cite{deepaugment2021,calian2021}, or coalescing multiple techniques~\cite{augmax2021}, and thus achieve good performance on different benchmarks. This strategy, though, leaves a big range of unintuitive design choices, making it hard to pinpoint which elements of these methods meaningfully contribute to the overall robustness. Meanwhile, the high complexity of recent methods \cite{augmax2021,calian2021} makes them impractical for large-scale tasks. Whereas, some methods are tailored to particular datasets and might not be general enough. Nonetheless, the problem of building robust classifiers is far from completely solved, and the gap between robust and standard accuracy is still large.

% To overcome this issue, the research community has favoured increasing the ``diversity'' of the training samples via data augmentation~\cite{autoaugment2019,augmix2020,deepaugment2021}. Intuitively, showing very diverse augmentations of an image to a network would increase the chance that the latter becomes invariant to some common corruptions. Still, 
% guaranteeing good coverage over the whole space of corruptions is hard. 
% {\color{blue}covering the full space of common corruptions is hard.}
%
% Hence, current literature relies mostly on increasing both computational and conceptual complexity \pf{on does not rely on increasing complexity... rephrase}, e.g., introducing DNNs for generating varied augmentations~\cite{deepaugment2021,calian2021}, or coalescing multiple techniques~\cite{augmax2021}, to increase the diversity of augmentations and achieve good performance on different benchmarks.
%
% Yet, the problem of building robust classifiers is far from completely solved, and the gap between robustness and standard accuracy is still large. {\color{blue}Besides, it is unclear if improving robustness requires sacrificing standard accuracy~\cite{tsiprasRobustnessMayBe2018}.}

\begin{figure}[t]
    \centering
    \includegraphics[width=0.7\columnwidth]{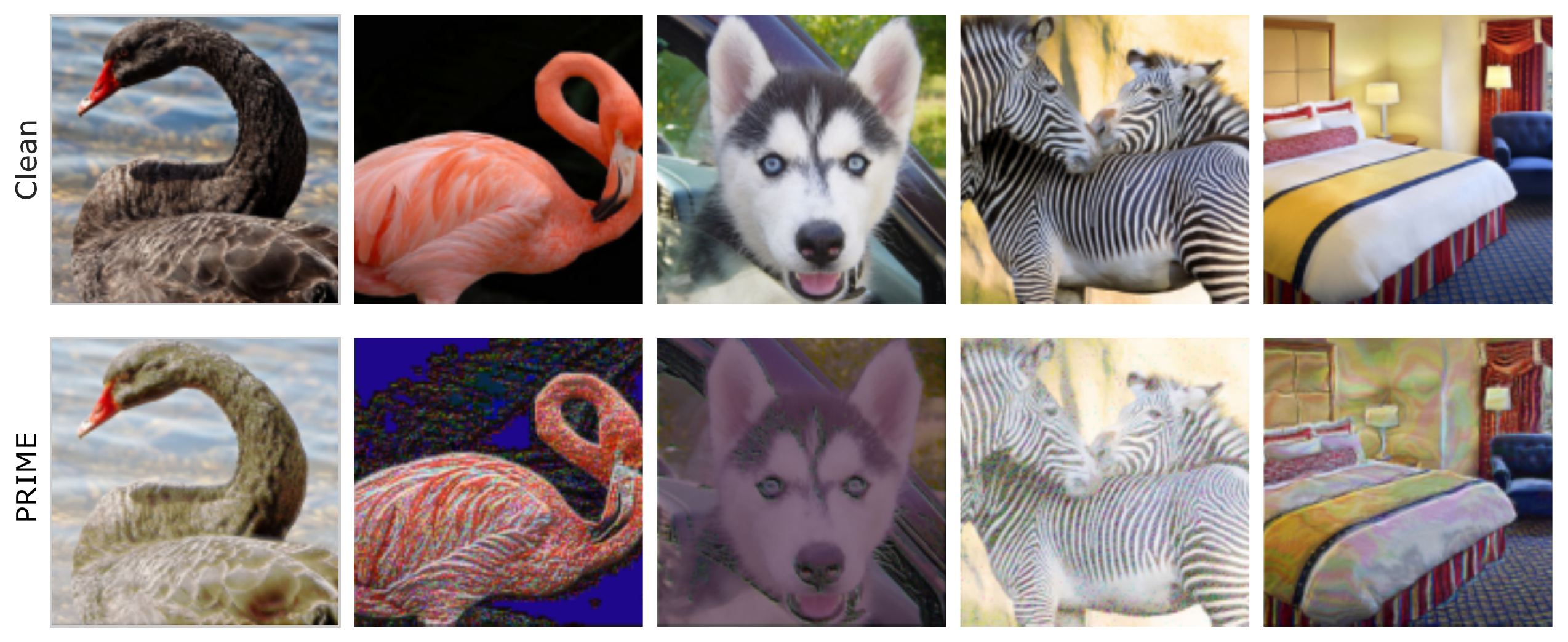}
    \caption{Images generated with PRIME, a simple method that uses a family of max-entropy transformations in different visual domains to create diverse augmentations.}
    \vspace*{-2mm}
    \label{fig:BaseMix_example}
\end{figure}

In this work, we take a step back and provide a systematic way for designing a simple, yet effective data augmentation scheme. By focusing on first principles, we formulate a new mathematical model for semantically-preserving corruptions, and build on basic concepts to characterize the notions of transformation strength and diversity using a few transformation primitives. Relying on this model, we propose \emph{PRIME}, a data augmentation scheme that draws transformations from a max-entropy distribution to efficiently sample from a large space of possible distortions (see~\cref{fig:BaseMix_example}). The performance of PRIME, alone, already tops the current baselines on different common corruption datasets, whilst it can also be combined with other methods to further boost their performance. Moreover, the simplicity and flexibility of PRIME allows to easily understand how each of its components contributes to improving robustness.

\noindent Altogether, the main contributions of our work include:
\begin{itemize}
    \item[$\bullet$] We introduce PRIME, a simple method that is built on a few guiding principles, which efficiently boosts robustness to common corruptions.
    \item[$\bullet$] We experimentally show that PRIME, despite its simplicity, achieves state-of-the-art robustness on multiple corruption benchmarks.  
    \item[$\bullet$] Last, our thorough ablation study sheds light on the necessity of having diverse transformations, on the role of mixing in the success of current methods, on the potential robustness-accuracy trade-off, and on the importance of online augmentations.
\end{itemize}

Overall, PRIME is a simple model-based scheme that can be easily understood, ablated, and tuned. Our work is an important step in the race for robustness against common corruptions, and we believe that it has the potential to become the new baseline for learning robust classifiers.

%%%%%%%%%%%%%%%%%%%%%%%%%%%%%%%%%%%%%%
\section{General model of visual corruptions}
\label{sec:model_of_cc}
%%%%%%%%%%%%%%%%%%%%%%%%%%%%%%%%%%%%%% 

In this work, motivated by the ``semantically-preserving'' nature of common corruptions, we define a new model of typical distortions. Specifically, we leverage the long tradition of image processing in developing techniques to manipulate images while retaining their semantics and construct a principled framework to characterize a large space of visual corruptions.

Let $\bm x:[0,1]^2\to[0,1]^3$ be a continuous image\footnote{In practice, we will work with discrete images on a regular grid.} mapping pixel coordinates $\bm r=(r_1, r_2)$ to RGB values. We define our model of common corruptions as the action on $\bm x$ of the following additive subgroup of the near-ring of transformations~\cite{near-ring}
\begin{equation}
    \mathcal{T}_{\bm x}=\left\{\sum_{i=1}^n \lambda_i\; g^i_1\circ \dots\circ g^i_m(\bm x): \:g^i_j\in\{\omega, \tau, \gamma\}, \lambda_i \in \R\right\},
\label{eq:cc_model}
\end{equation}
where $\omega, \tau$ and $\gamma$ are random primitive transformations which distort $\bm x$ along the spectral ($\omega$), spatial ($\tau$), and color ($\gamma$) domains. As we will see, defining each of these primitives in a principled and coherent fashion will be enough to construct a set of perturbations which covers most types of visual corruptions.

% All our primitives are designed based on two guiding principles: maximum entropy and smoothness. The principle of maximum entropy~\cite{cover_info} suggests using distributions of functions which are as unbiased as possible given their constraints. Although identifying formally the whole space of common corruptions is hard, the principle of maximum entropy guarantees a good coverage over each of our primitives. That is, with each new augmentation we provide the network with the maximum additional information possible about the domain of the transformation. In~\cref{app:max-entropy}, we provide a formal derivation of each maximum entropy distribution. The principle of smoothness, on the other hand, formalizes the notion of physical plausibility, as most naturally occurring processes are smooth.

To guarantee as much diversity as possible in our model, we follow the principle of maximum entropy to define our distributions of transformations~\cite{cover_info}. Note that using a set of augmentations that guarantees maximum entropy comes naturally when trying to optimize the sample complexity derived from certain information-theoretic generalization bounds, both in the clean~\cite{xuInfoBound} and corrupted settings~\cite{OODInfoBound}. Specifically, the principle of maximum entropy postulates favoring those distributions that are as unbiased as possible given the set of constraints that define a family of distributions. In our case, these constraints are given in the form of an expected strength $\sigma^2$, some boundary conditions, e.g., the displacement field must be zero at the borders of an image, and finally the desired smoothness level $K$. The principle of smoothness helps formalize the notion of physical plausibility, as most naturally occurring processes are smooth.

Formally, let $\mathcal{I}$ denote the space of all images, and let $f:\mathcal{I}\to \mathcal{I}$ be a random image transformation distributed according to the law $\mu$. Further, let us define a set of constraints $\mathcal{C}\subseteq \mathcal{F}$, which restricts the domain of applicability of $f$, i.e., $f\in\mathcal{C}$, and where $\mathcal{F}$ denotes the space of functions $\mathcal{I}\to \mathcal{I}$. The principle of maximum entropy postulates using the distribution $\mu$ which has maximum entropy given the constraints:
\begin{align}
    \underset{\mu}{\text{maximize}}\quad &  H(\mu)=-\int_{\mathcal{F}}\, \mathrm{d}\mu(f) \log(\mu(f)) \label{eq:max_entropy}\\
    \text{subject to}\quad & f\in\mathcal C\quad \forall f\in\operatorname{supp}(\mu), \nonumber
\end{align}
where $H(\mu)$ represents the entropy of the distribution $\mu$~\cite{cover_info}. In its general form, solving \cref{eq:max_entropy} for any set of constraints $\mathcal{C}$ is intractable. In \cref{app:max-entropy}, we formally derive the analytical expressions for the distributions of each of our family of transformations, by leveraging results from statistical physics~\cite{beale}.

In what follows, we describe the analytical solutions to \cref{eq:max_entropy} for each of our basic primitives. In general, these distributions are governed by two parameters: $K$ to control smoothness, and $\sigma^2$ to control strength. These transformations fall back to identity mappings when $\sigma^2=0$, independently of $K$. 

\smallskip\noindent\textbf{Spectral domain} We parameterize the distribution of random spectral transformations using random filters $\bm\omega(\bm r)$, such that the transformation output follows
\begin{equation}
    \omega(\bm x)(\bm r)=\left(\bm x * \left(\bm\delta+\bm \omega'\right)\right)(\bm r),
\label{eq:spectral_domain}
\end{equation}
where, $*$ is the convolution operator,  $\bm\delta(\bm r)$ represents a Dirac delta, i.e., identity filter, and $\bm\omega'(\bm r)$ is implemented in the discrete grid as an FIR filter of size $K_\omega\times K_\omega$ with i.i.d random entries distributed according to $\mathcal{N}(0, \sigma^2_\omega)$. Here, $\sigma^2_\omega$ governs the transformation strength, while larger $K_\omega$ yields filters of higher spectral resolution. The bias $\bm\delta(\bm r)$ retains the output close to the original image.

\smallskip\noindent\textbf{Spatial domain} We model our distribution of random spatial transformations, which apply random perturbations over the coordinates of an image, as
\begin{equation}
    \tau(\bm x)(\bm r) = \bm x(\bm r + \bm\tau'(\bm r)).
\end{equation}
This model has been recently proposed in~\cite{diffeo} to define a distribution of random smooth diffeomorphisms in order to study the stability of neural networks to small spatial transformations. To guarantee smoothness but preserve maximum entropy, the authors propose to parameterize the vector field $\bm\tau'$ as
\begin{equation}
    \bm\tau'(\bm r)=\sum_{i^2+j^2\leq K^2_\tau}\beta_{i,j}\sin(\pi i \bm r_1)\sin(\pi j \bm r_2), \label{eq:spatial_transform}
\end{equation}
where $\beta_{i,j}\sim\mathcal{N}(0,\sfrac{\sigma^2_\tau}{(i^2+j^2)})$. Such choice guarantees that the resulting mapping is smooth according to the cut frequency $K_\tau$, while $\sigma^2_\tau$ determines its strength.

\smallskip\noindent\textbf{Color domain} Following a similar approach, we define the distribution of random color transformations as random mappings $\gamma$ between color spaces
\begin{equation}
    \gamma(\bm x)(\bm r) = \bm x(\bm r) + \sum_{n=0}^{K_\gamma} \bm\beta_n \odot\sin\left(\pi n \,\bm x(\bm r)\right),\label{eq:color_transform}
\end{equation}
where $\bm{\beta}_n\sim\mathcal{N}(0, \sigma^2_\gamma\bm I_3)$, with $\odot$ denoting elementwise multiplication. Again, $K_\gamma$ controls the smoothness of the transformations and $\sigma^2_\gamma$ their strength. Compared to~\cref{eq:spatial_transform}, the coefficients in~\cref{eq:color_transform} are not weighted by the inverse of the frequency, and have constant variance. In practice, we observe that reducing the variance of the coefficients for higher frequencies creates color mappings that are too smooth and almost imperceptible, so we decided to drop this dependency. 

\AlgoDontDisplayBlockMarkers
\RestyleAlgo{ruled}
\SetAlgoNoLine
\LinesNumbered
\begin{algorithm}[t]
    \footnotesize
    \algnewcommand{\LeftComment}[1]{\Statex \(\triangleright\) #1}
 	\KwIn{Image $\bm x$, primitives $\mathcal{G}=\{\operatorname{Id},\omega, \tau\, \gamma\}$, where $\operatorname{Id}$ is the identity operator} 
 	\KwOut{Augmented image $\tilde{\bm x}$}
 	\BlankLine
    $\tilde{\bm x}_0 \gets \bm x$\\
	\For{$i\in \{1,\dots,n\}$}
	{
	    $\tilde{\bm x}_i \gets \bm x$\\
        \For{$j\in\{1,\dots,m\}$}
        {
            $g\sim\mathcal{U}(\mathcal{G})$ \Comment{Strength $\sigma\sim\mathcal{U}(\sigma_{\text{min}}, \sigma_{\text{max}})$} \\
            $\tilde{\bm x}_i \gets g(\tilde{\bm x}_i)$
        }
    }
    $\bm \lambda\sim\operatorname{Dir}(\bm 1)$ \Comment{Random Dirichlet convex coefficients}\\
    $\tilde{\bm x}\gets \sum_{i=0}^n \lambda_i \tilde{\bm x}_i$ \\
%  	\KwRet{$\tilde{\bm x}$}
\caption{PRIME}
\label{alg:prime}
\end{algorithm}

Finally, we note that PRIME is very flexible with respect to its core primitives. In particular, PRIME can be easily extended to include other distributions of maximum entropy transformations that suit an objective task. For example, one might add the distribution of maximum entropy additive perturbations given by $\eta(\bm x)(\bm r)= \bm x(\bm r)+\bm\eta'(\bm r)$, where $\bm \eta'(\bm r)\sim\mathcal{N}(0, \sigma^2_\eta)$. Nonetheless, since most benchmarks of visual corruptions disallow the use of additive perturbations during training~\cite{corruptions2019}, we do not include an additive perturbation category.

% since most benchmarks of visual corruptions disallow the use of additive perturbations during training~\cite{corruptions2019}, our model does not include an additive perturbation category. Nevertheless, for the sake of completeness, and depending on the application, one may also want to define an extra set of max-entropy transformations given by $\eta(\bm x)(\bm r)= \bm x(\bm r)+\bm\eta'(\bm r)$, where $\bm \eta'(\bm r)\sim\mathcal{N}(0, \sigma^2_\eta)$, and which could be readily integrated in~\cref{eq:cc_model}.

Overall, as demonstrated by our results in \cref{subsec:performance_results,subsec:role_of_mixing}, our model is very flexible and can cover a large part of the semantic-preserving distortions. It also allows to easily control the strength and style of the transformations with just a few parameters. Moreover, changing the transformation strength enables to control the trade-off between corruption robustness and standard accuracy, as shown in Sec.~\ref{subsec:robustness-accuracy-tradeoff}. In what follows, we use this model to design an efficient augmentation scheme to build classifiers robust to common corruptions.

% Overall, as we will see, our model is very flexible and can cover a large part of the semantic-preserving distortions. 
% % \pf{do we actually show that anywhere?}. 
% It also allows to easily control the strength and style of the transformations with just a few parameters, and also provides a systematic way to control the trade-off between robustness and standard accuracy. In what follows, we use this model to design an efficient augmentation scheme to build classifiers robust to common corruptions.

%%%%%%%%%%%%%%%%%%%%%%%%%%%%%%%%%%%%%%
\section{PRIME: A simple augmentation scheme}
\label{sec:prime}
%%%%%%%%%%%%%%%%%%%%%%%%%%%%%%%%%%%%%%

We now introduce PRIME, a simple yet efficient augmentation scheme that uses our \textbf{PRI}mitives of \textbf{M}aximum \textbf{E}ntropy to confer robustness against common corruptions. The pseudo-code of PRIME is given in~\cref{alg:prime}, which draws a random sample from~\cref{eq:cc_model} using a convex combination of a composition of basic primitives. Below we describe the main implementation details.
% of our algorithm.

\begin{figure}[t]
    \centering
    \includegraphics[width=0.65\columnwidth]{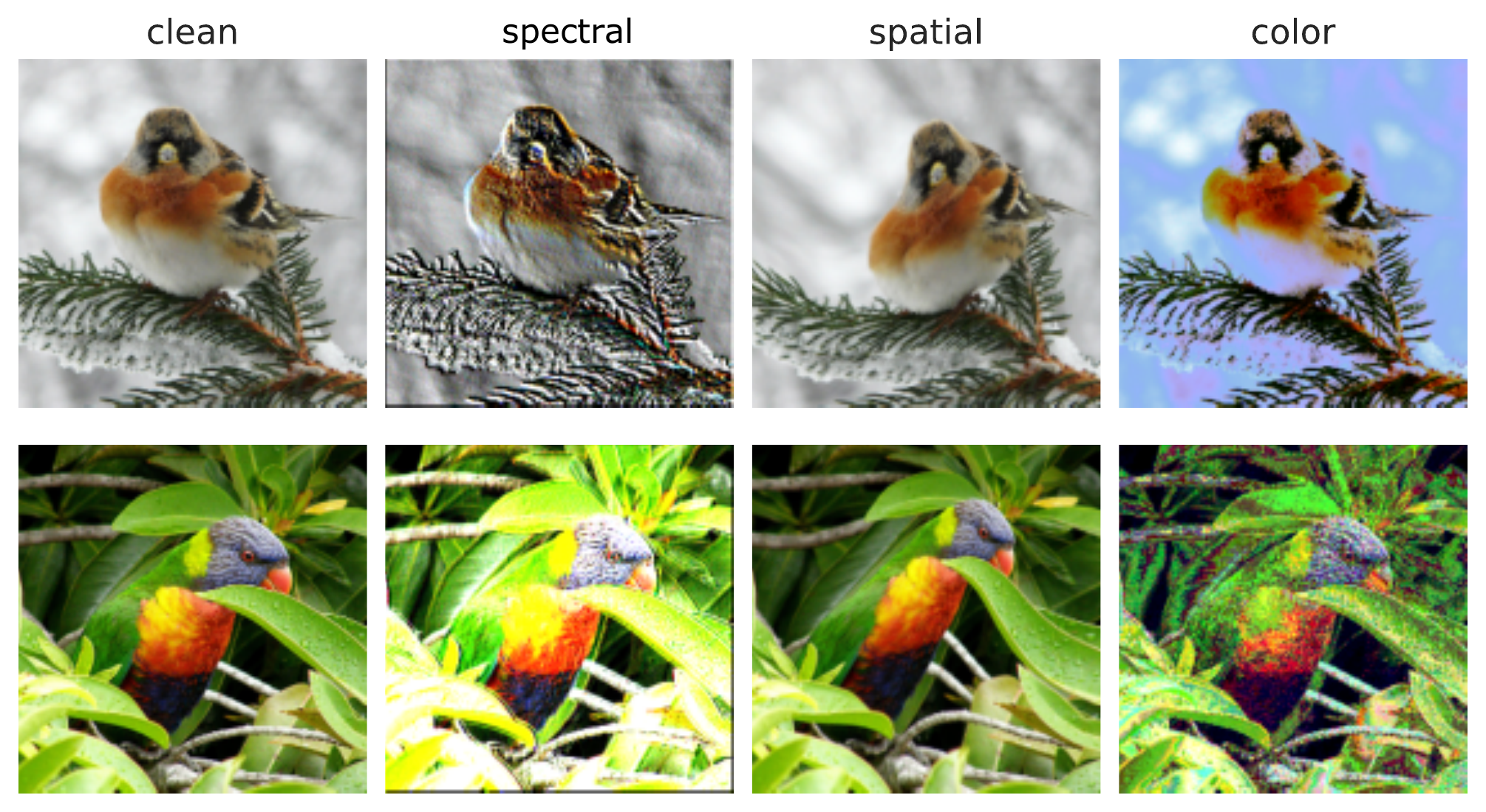}
    \caption{Images generated with the transformations of our common corruptions model. Despite the perceptibility of the distortion, the image semantics are preserved.}
    \vspace*{-2mm}
    \label{fig:visual_inspection}
\end{figure}

\smallskip\noindent\textbf{Parameter selection}
It is important to ensure that the semantic information of an image is preserved after it goes through PRIME. As measuring semantic preservation quantitatively is not simple, we subjectively select each primitive's parameters based on visual inspection, ensuring maximum permissible distortion while retaining the semantic content of the image. However, to avoid relying on a specific strength for each transformation, PRIME stochastically generates augmentations of different strengths by sampling $\sigma$ from a uniform distribution, with different minimum and maximum values for each primitive. \Cref{fig:visual_inspection} shows some visual examples for each kind of transformation, while additional visual examples along with the details of all the parameters can be found in~\cref{app:implementation_details}. 

For the color primitive, we observed that fairly large values for $K_\gamma$ (in the order of $500$) are important for covering a large space of visual distortions. Unfortunately, implementing such a transformation can be memory inefficient. To avoid this issue, PRIME uses a slight modification of \cref{eq:color_transform} and combines a fixed number $\Delta$ of consecutive frequencies randomly chosen in the range $[0, K_\gamma]$. 
% Finally, as some of our transformations can push the images outside of their color range, we always clip the output of each transformation so that it lies on $[0,1]^3$.

\smallskip\noindent\textbf{Mixing transformations}
The concept of mixing has been a recurring theme in the augmentation literature~\cite{mixup2018,cutmix2019,augmix2020,augmax2021} and PRIME follows the same trend. In particular, \cref{alg:prime} uses a convex combination of $n$ basic augmentations consisting of the composition of $m$ of our primitive transformations. In general, the convex mixing procedure (i) broadens the set of possible training augmentations, and (ii) ensures that the augmented image stay close to the original one. We later provide empirical results which underline the efficacy of mixing in~\cref{subsec:role_of_mixing}. Overall, the exact mixing parameters are provided in~\cref{app:implementation_details}. Note that, the basic skeleton of PRIME is similar to that of AugMix. However, as we will see next, incorporating our maximum entropy transformations leads to significant gains in common corruptions robustness over AugMix. 

% Note that, algorithmically, PRIME and AugMix~\cite{augmix2020} are identical except for the choice of transformation primitives, and have a similar computational complexity (e.g., on ImageNet-100, AugMix takes 255 and PRIME takes 269 minutes for training). However, as we will see next, incorporating our maximum entropy transformations significantly boosts robustness to common corruptions. \pf{are there any other component that make both schemes different? that could help outlining novetly? It looks a bit on the weak side, if written like in this paragraph.}

%%%%%%%%%%%%%%%%%%%%%%%%%%%%%%%%%%%%%%
\section{Performance analysis}
\label{sec:experimental}
%%%%%%%%%%%%%%%%%%%%%%%%%%%%%%%%%%%%%%

In this section, we compare the classification performance of our method on multiple datasets with that of two current approaches: AugMix and DeepAugment (DA). We illustrate that PRIME significantly advances the corruption robustness over that of AugMix and DeepAugment on all the benchmarks. We also show that our method yields additional benefits when employed in concert with unsupervised domain adaptation~\cite{bnadapt_bethge2021}.

\begin{table}[t]
    \centering
    \footnotesize
    \caption{Clean and corruption accuracy, and mean corruption error (mCE) for different methods with ResNet-18 on C-10, C-100, IN-100 and ResNet-50 on IN. mCE is the mean corruption error on common corruptions un-normalized for C-10 and C-100; normalized relative to standard model on IN-100 and IN. $^\dagger$ indicates that JSD consistency loss is not used. $^*$Models taken from~\cite{robustbench2021}.}
    \begin{tabular}{clccc}
        \toprule
        \multirow{2}{*}{Dataset} & \multicolumn{1}{c}{\multirow{2}{*}{Method}} & Clean & \multicolumn{2}{c}{Common Corruption}\\
        & & Acc ($\uparrow$) & Acc ($\uparrow$) & mCE ($\downarrow$)\\
        \midrule
        \multirow{3}{*}{C-10} & Standard & 95.0 & 74.0 & 24.0\\
        & AugMix & 95.2 & 88.6 & 11.4\\
        & PRIME & 94.2 & \textbf{89.8} & \textbf{10.2}\\
        \midrule
        \multirow{3}{*}{C-100}& Standard & 76.7 & 51.9 & 48.1\\
        & AugMix & 78.2 & 64.9 & 35.1\\
        & PRIME & 78.4 & \textbf{68.2} & \textbf{31.8}\\
        \midrule
        \multirow{6}{*}{IN-100} & Standard & 88.0 & 49.7 & 100.0\\
        & AugMix & 88.7	& 60.7 & 79.1\\
        & DA & 86.3 & 67.7 & 68.1\\
        & PRIME & 85.9 & \textbf{71.6} & \textbf{61.0}\\
        \cmidrule{2-5}
        & DA\texttt{+}AugMix & 86.5 & 73.1 & 57.3\\
        & DA\texttt{+}PRIME & 84.9 & \textbf{74.9} & \textbf{54.6}\\
        \midrule
        \multirow{7}{*}{IN} & Standard$^*$ & 76.1 & 38.1 & 76.7\\
        & AugMix$^*$ & 77.5 & 48.3 & 65.3 \\
        & DA$^*$ & 76.7 & 52.6 & 60.4 \\
        & PRIME$^\dagger$ & 77.0 & \textbf{55.0} & \textbf{57.5}\\
        % & (w/o JSD) & &\\
        \cmidrule{2-5}
        & DA\texttt{+}AugMix & 75.8 & 58.1 & 53.6\\
        & DA\texttt{+}PRIME$^\dagger$ & 75.5 & \textbf{59.9} & \textbf{51.3}\\
        % & (w/o JSD) & &\\ 
        \bottomrule
    \end{tabular}
    \vspace*{-2mm}
    \label{tab:results-main}
\end{table}

\subsection{Training setup}
\label{subsec:training_setup}

We consider the CIFAR-10 (C-10), CIFAR-100 (C-100)~\cite{cifar102009}, ImageNet-100 (IN-100) and ImageNet (IN)~\cite{imagenet2009} datasets. IN-100 is a $100$-class subset of IN obtained by selecting every $10$\textsuperscript{th} class in WordNet ID order. We train a ResNet-18~\cite{resnet2016} on C-10, C-100 and IN-100; and a ResNet-50 on IN for $100$ epochs. Following AugMix, and for a complete comparison, we also integrate the Jensen-Shannon divergence (JSD)-based consistency loss in PRIME which compels the network to learn similar representations for differently augmented versions of the same input image. Detailed training setup appears in~\cref{app:exp-setup}. We evaluate our trained models on the common corrupted versions (C-10-C, C-100-C, IN-100-C, IN-C) of the aforementioned datasets. The common corruptions \cite{corruptions2019} constitute $15$ image distortions each applied with $5$ different severity levels. These corruptions can be grouped into four categories, viz. noise, blur, weather and digital.

\subsection{Robustness to common corruptions}
\label{subsec:performance_results}

% \begin{table}[t]
%     \centering
%     \footnotesize
%     \caption{Corruption and clean accuracy (in parenthesis) for different methods with ResNet-18 on C-10, C-100, IN-100 and ResNet-50 on IN. $^\dagger$ indicates that JSD consistency loss is not used. $^*$Models taken from \texttt{RobustBench}~\cite{robustbench2021}.}
%     \aboverulesep=0ex
%     \belowrulesep=0ex
%     \tabcolsep=8pt
%     \begin{tabular}{lcccc}
%         \toprule
%         Method & C-10 & C-100 & IN-100 & IN \\
%         \midrule
%         Standard & 74.0 (95.0) & 51.9 (76.7) & 49.7 (88.0) & 38.1 (76.1)$^*$ \\
%         AugMix & 88.6 (95.2) & 64.9 (78.2) & 60.7 (88.7) & 48.3 (77.5)$^*$ \\
%         PRIME &  \textbf{89.0} (93.1) & \textbf{68.3} (77.6) & \textbf{71.6} (85.9) & \textbf{55.0} (77.0)$^\dagger$ \\
%         \cmidrule{1-5}
%         DA & -- & -- & 67.7 (86.3) & 52.6 (76.7)$^*$ \\
%         DA\texttt{+}AugMix & -- & -- & 73.1 (86.5) & 58.1 (75.8) \\
%         DA\texttt{+}PRIME & -- & -- & \textbf{74.9} (84.9) & \textbf{59.9} (75.5)$^\dagger$\\
%         \bottomrule
%     \end{tabular}
%     \vspace*{-2mm}
%     \label{tab:results-main}
% \end{table}

In order to assess the effectiveness of PRIME, we evaluate its performance against C-10, C-100, IN-100 and IN common corruptions. The results are summarized in~\cref{tab:results-main}\footnote{We provide the per-corruption performance of every method in~\cref{app:performance-per-corruption}.}. Amongst individual methods, PRIME yields superior results compared to those obtained by AugMix and DeepAugment alone and advances the baseline performance on the corrupted counterparts of the four datasets. As listed, PRIME pushes the corruption accuracy by $1.2\%$ and $3.3\%$ on C-10-C and C-100-C respectively over AugMix. On IN-100-C, a more complicated dataset, we observe significant improvements wherein PRIME outperforms AugMix by $10.9\%$. In fact, this increase in performance hints that our primitive transformations are actually able to cover a larger space of image corruptions, compared to the restricted set of AugMix. Interestingly, the random transformations in PRIME also lead to a $3.9\%$ boost in corruptions accuracy over DeepAugment despite the fact that DeepAugment leverages additional knowledge to augment the training data via its use of pre-trained architectures. Moreover, PRIME provides cumulative gains when combined with DeepAugment, reducing the mean corruption error (mCE) of prior art (DA\texttt{+}AugMix) by $2.7\%$ on IN-100-C. Lastly, we also evaluate the performance of PRIME on full IN-C. However, we do not use JSD in order to reduce computational complexity. Yet, even without the JSD loss, PRIME outperforms, in terms of corruption accuracy, both AugMix (with JSD) and DeepAugment by $6.7\%$ and $2.4\%$ respectively, while the mCE is reduced by $7.8\%$ and $2.9\%$. And last, when PRIME is combined with DeepAugment, it also surpasses the performance of DA\texttt{+}AugMix (with JSD), reaching a corruption accuracy of almost $60\%$ and an mCE of $51.3\%$. Note here, that, not only PRIME achieves superior robustness, but it does so efficiently. Compared to standard training on IN-100, AugMix requires 1.20x time and PRIME requires 1.27x. In contrast, DA is tedious and we do not measure its runtime since it also requires the training of two large image-to-image networks for producing augmentations, and can only be applied offline.

\subsection{Unsupervised domain adaptation}
\label{subsec:unsupervised-adaptation}

\begin{table}[ht]
    \centering
    \footnotesize
    \caption{Performance of different methods in concert with domain adaptation on IN-100. Partial adaptation uses $8$ samples; full adaptation uses $400$ corrupted samples. Network used: ResNet-18.}
    \begin{tabular}{lccccc}
        \toprule
        & \multicolumn{4}{c}{IN-100-C acc. ($\uparrow$)} & IN-100 ($\uparrow$)\\
        \cmidrule{2-6}
        Method & w/o & single & partial & full & single\\
        \midrule
        Standard & 49.7 & 53.8 & 62.0 & 63.9 & 88.1\\
        AugMix & 60.7 & 65.5 & 71.3 & 73.0 & \textbf{88.3}\\
        DA & 67.7 & 70.2 & 72.7 & 74.6 & 86.3\\
        PRIME & \textbf{71.6} & \textbf{73.5} & \textbf{75.3} & \textbf{76.6} & 85.7\\
        \bottomrule
    \end{tabular}
    \vspace*{-2mm}
    \label{tab:bn-adapt}
\end{table}

Recently, robustness to common corruptions has also been of significant interest in the field of unsupervised domain adaptation~\cite{bnadapt_so2021,bnadapt_bethge2021}. The main difference is that, in domain adaptation, one exploits the limited access to test-time corrupted samples to adjust certain network parameters. Hence, it would be interesting to investigate the utility of PRIME under the setting of domain adaption. 

To that end, we combine our method with the adaption trick of~\cite{bnadapt_bethge2021}. Specifically, we adjust the batch normalization (BN) statistics of our models using a few corrupted samples. Suppose $z_s \in \{\mu_s$, $\sigma_s\}$ are the BN mean and variance estimated from the training data, and $z_t \in \{\mu_t$, $\sigma_t\}$ are the corresponding statistics computed from $n$ unlabelled, corrupted test samples, then we re-estimate the BN statistics as follows.
\begin{equation}
    \hat{z} = \frac{N}{N+n}z_s + \frac{n}{N+n}z_t.
\end{equation}
We consider three adaptation scenarios: single sample ($n=1, N=16$), partial ($n=8, N=16$) and full ($n=400, N=0$) adaptation. Here, we do not perform parameter tuning for $N$. As shown in \cref{tab:bn-adapt}, simply correcting BN statistics using as little as $8$ corrupted samples pushes the corruption accuracy of PRIME from $71.6\%$ to $75.3\%$. In general, PRIME yields cumulative gains in combination with adaptation and has the best IN-100-C accuracy.

%%%%%%%%%%%%%%%%%%%%%%%%%%%%%%%%%%%%%%
\section{Robustness insights using PRIME}
\label{sec:analysis}
%%%%%%%%%%%%%%%%%%%%%%%%%%%%%%%%%%%%%%

In this section, we exploit the simplicity and the controllable nature of PRIME to investigate different aspects behind robustness to common corruptions. We first analyze how each transformation domain contributes to the overall robustness of the network. Then, we empirically locate and justify the benefits of mixing the transformations of each domain. Moreover, we demonstrate the existence of a robustness-accuracy trade-off, and, finally, we comment on the low-complexity benefits of PRIME in different data augmentation settings.

\subsection{Contribution of transformations}
\label{sec:orthogonality}

\begin{table}[t]
    \centering
    \footnotesize
    \caption{Impact of the different max-entropy primitives ($\omega$: spectral, $\gamma$: color, $\tau$: spatial) in PRIME on common corruption accuracy ($\uparrow$) of a ResNet-18. All the transformations are essential for the performance of PRIME. The JSD loss is \emph{not} used.}
    \begin{tabular}{lcccccc}
        \toprule
        {Transform} & {IN-100-C} & {Noise} & {Blur} & {Weather} & {Digital} & {IN-100}\\
        \midrule
        None & 49.7 & 27.3 & 48.6 & 54.8 & 62.6 & \textbf{88.0}\\
        $\omega$ & 64.1 & 60.7 & 55.4 & 66.6 & 72.9 & 87.3\\
        $\tau$ & 53.8 & 30.1 & 56.2 & 57.6 & 65.4 & 87.0\\
        $\gamma$ & 59.9 & 67.4 & 52.6 & 54.4 & 67.1 & 86.9\\
        $\omega$\texttt{+}$\tau$ & 64.5 & 58.5 & 57.3 & \textbf{66.8} & 73.9 & 87.7\\
        $\omega$\texttt{+}$\gamma$ & 67.5 & 77.2 & 55.7 & 65.3 & 74.2 & 87.1\\
        $\tau$\texttt{+}$\gamma$ & 63.3 & 74.7 & 57.4 & 56.2 & 67.8 & 86.2\\
        $\omega$\texttt{+}$\tau$\texttt{+}$\gamma$ & \textbf{68.8} & \textbf{78.8} & \textbf{58.3} & 66.0 & \textbf{74.8} & 87.1\\
        \bottomrule
    \end{tabular}
    \vspace*{-2mm}
    \label{tab:orthogonality-in100}
\end{table}

We want to understand how the transformations in each domain of~\cref{eq:cc_model} contribute to the overall robustness. To that end, we conduct an ablation study on IN-100-C by training a ResNet-18 with the max-entropy transformations of PRIME individually or in combination. As shown in~\cref{tab:orthogonality-in100}, spectral transformations mainly help against blur, weather and digital corruptions. Spatial operations also improve on blurs, but on elastic transforms as well (digital). On the contrary, color transformations excel on noises and certain high frequency digital distortions, e.g., pixelate and JPEG artefacts, and have minor effect on weather changes. Besides, incrementally combining the transformations lead to cumulative gains e.g., spatial\texttt{+}color help on both noises and blurs. Yet, for obtaining the best results, the combination of all transformations is required. This means that each transformation increases the coverage over the space of possible distortions and the increase in robustness comes from their cumulative contribution.

\subsection{The role of mixing}
\label{subsec:role_of_mixing}

\begin{figure}[t]
    \centering
    \footnotesize
    \begin{minipage}[c]{0.445\columnwidth}
        \centering
        \includegraphics[width=\linewidth]{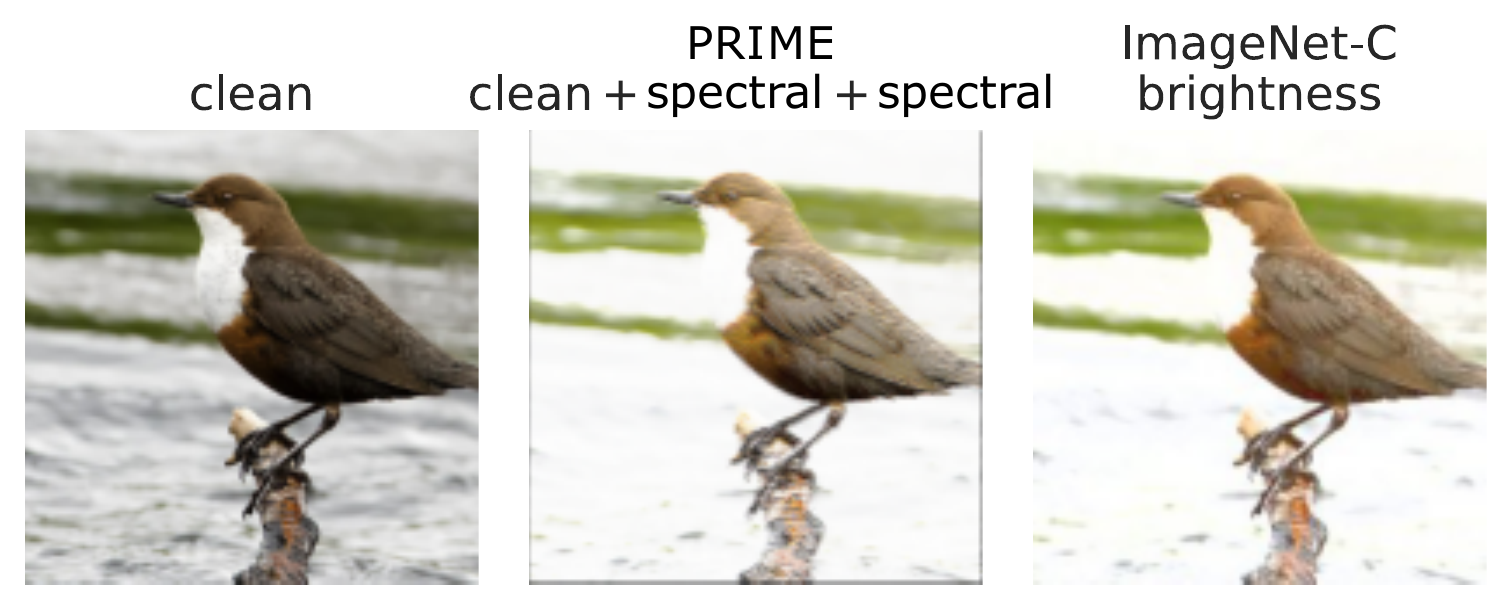}
        % \captionsetup{justification=centering}
        \vspace*{-2mm}
        % \caption{$\texttt{clean+}\texttt{spectral}\texttt{+spectral}\approx\texttt{brightness}$}
        \label{fig:mixing-examples:a}
    \end{minipage}
    \begin{minipage}[c]{0.445\columnwidth}
        \centering
        \includegraphics[width=\linewidth]{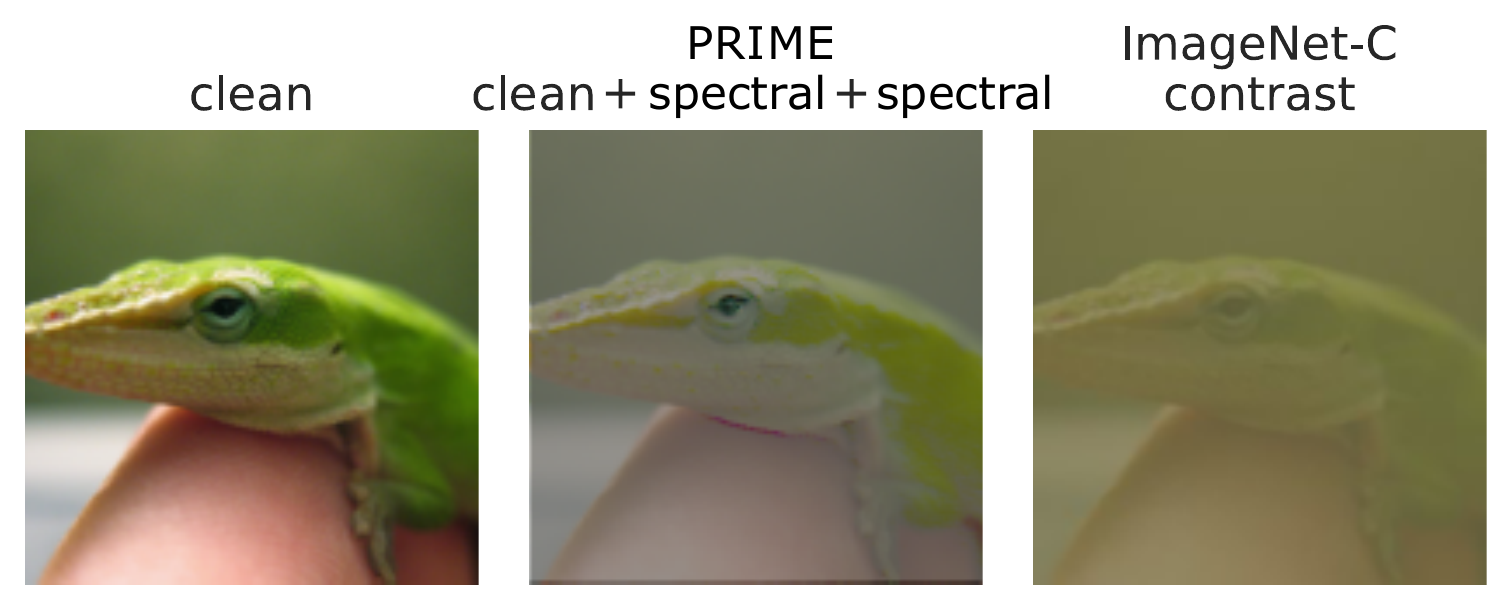}
        % \captionsetup{justification=centering}
        \vspace*{-2mm}
        % \caption{$\texttt{clean+}\texttt{spectral}$ \\$\texttt{+spectral}\approx\texttt{contrast}$}
        \label{fig:mixing-examples:b} 
    \end{minipage}
    \begin{minipage}[c]{0.445\columnwidth}
        \centering
        \includegraphics[width=\linewidth]{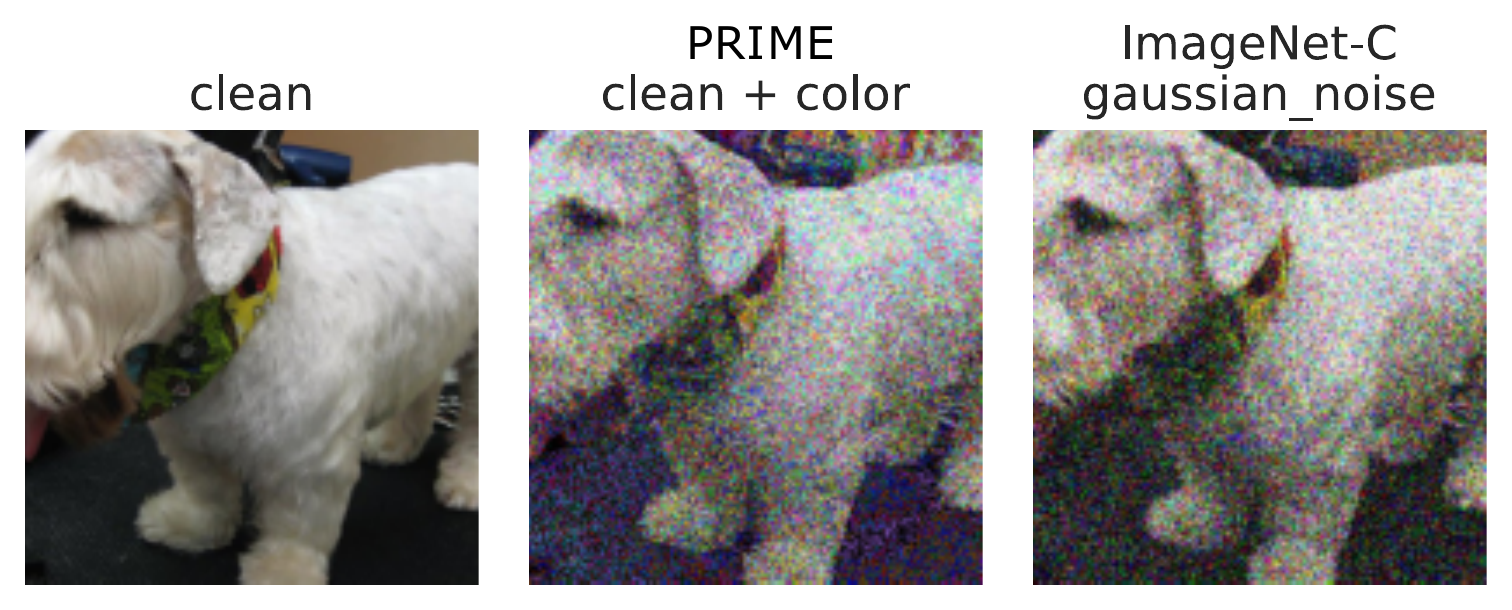}
        % \captionsetup{justification=centering}
        \vspace*{-4mm}
        % \caption{$\texttt{clean+}\texttt{color}$
        % \\$\approx\texttt{gaussian\_noise}$}
        \label{fig:mixing-examples:d}
    \end{minipage}
    \begin{minipage}[c]{0.445\columnwidth}
        \centering
        \includegraphics[width=\linewidth]{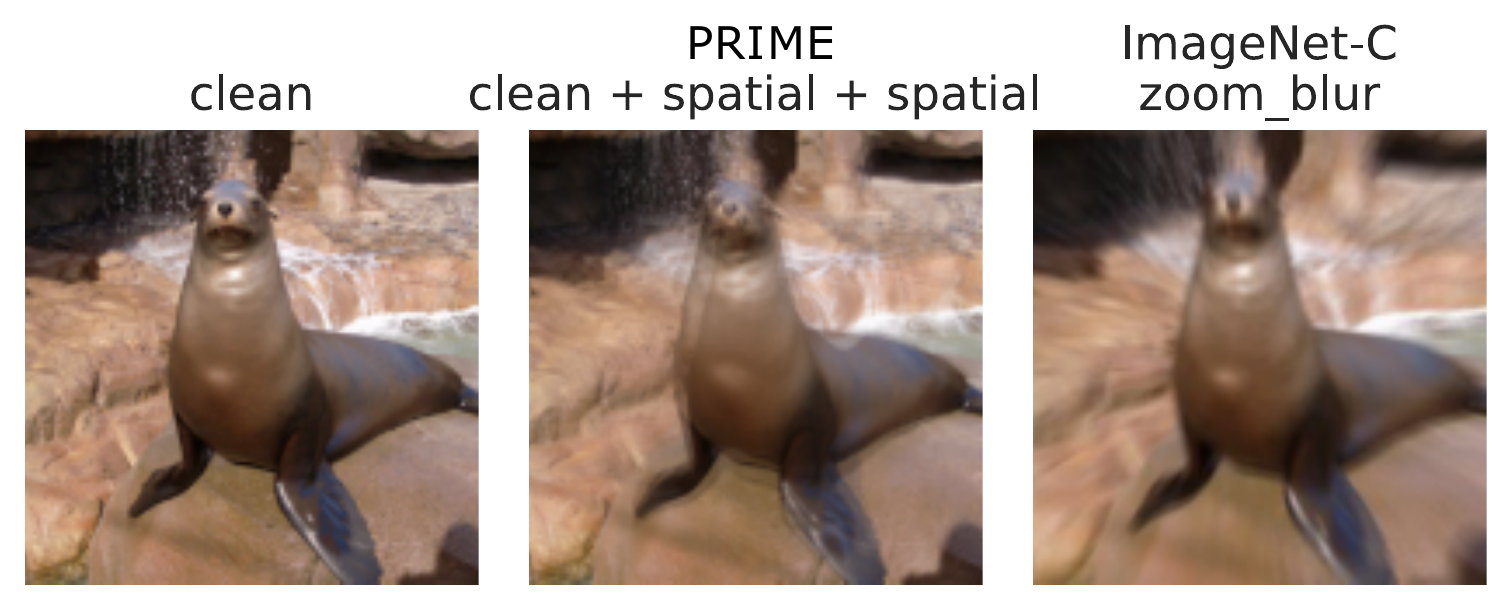}
        % \captionsetup{justification=centering}
        \vspace*{-4mm}
        % \caption{$\texttt{clean+}\texttt{spatial+}\texttt{spatial}$
        % \\$\approx\texttt{zoom\_blur}$}
        \label{fig:mixing-examples:e}
    \end{minipage}
    \caption{Mixing produces images that are visually similar to the test-time corruptions. Each example shows the clean image, the PRIME image and the corresponding common corruption that resembles the image produced by mixing. We also report the mixing combination used for recreating the corruption. See~\cref{app:mixing_examples} for additional examples.}
    \vspace*{-2mm}
    \label{fig:mixing-examples}
\end{figure}

% In most data augmentation methods, besides the importance of the transformations themselves, mixing has been claimed as an essential module for increasing diversity in the training process~\cite{mixup2018,cutmix2019,augmix2020,augmax2021}. In our attempt to provide insights on the role of mixing in the context of common corruptions, we found out that it is capable of constructing augmented images that look perceptually similar to their corrupted counterparts. In fact, the improvements on specific corruption types observed in~\cref{tab:orthogonality-in100} can be largely attributed to mixing. As exemplified in~\cref{fig:mixing-examples:a}~and~\cref{fig:mixing-examples:b}, careful combinations of spectral transformations with the clean image introduce brightness and contrast-like artefacts that look similar to the corresponding corruptions in IN-C. Also, combining spatial transformations creates blur-like artefacts that look identical to zoom blur in IN-C (\cref{fig:mixing-examples:e}). Finally, notice in~\cref{fig:mixing-examples:d} how mixing color transformations helps fabricate corruptions of the ``noise'' category. This means that the max-entropy color model of PRIME enables robustness to different types of noise without explicitly adding any during training. This might explain the significant improvement over the ``noise'' category in~\cref{tab:results-in100}.
In most data augmentation methods, besides the importance of the transformations themselves, mixing has been claimed as an essential module for increasing diversity in the training process~\cite{mixup2018,cutmix2019,augmix2020,augmax2021}. In our attempt to provide insights on the role of mixing in the context of common corruptions, we found out that it is capable of constructing augmented images that look perceptually similar to their corrupted counterparts. In fact, the improvements on specific corruption types observed in~\cref{tab:orthogonality-in100} can be largely attributed to mixing. As exemplified in~\cref{fig:mixing-examples}, careful combinations of spectral transformations with the clean image introduce brightness and contrast-like artefacts that look similar to the corresponding corruptions in IN-C. Also, combining spatial transformations creates blur-like artefacts that look identical to zoom blur in IN-C. Finally, notice how mixing color transformations helps fabricate corruptions of the ``noise'' category. This means that the max-entropy color model of PRIME enables robustness to different types of noise without explicitly adding any during training. 
% This might explain the significant improvement over the ``noise'' category in~Tab.~6 of Appendix H.

Note that one of the main goals of data augmentation is to achieve maximum coverage of the space of possible distortions using a limited transformation budget, i.e., within a few training epochs. The principle of max-entropy guarantees this within each primitive, but the effect of mixing on the overall space is harder to quantify. In this regard, we can use the distance in the embedding space, $\phi$, of a SimCLRv2~\cite{simclr} model as a proxy for visual similarity~\cite{lpips,moayeri_sample_efficient_iccv}. We are interested in measuring how mixing the base transformations changes the likelihood that an augmentation scheme generates some sample during training that is visually similar to some of the common corruptions. To that end, we randomly select $N=1000$ training images $\{\bm x_n\}_{n=1}^N$ from IN, along with their $C=75$ ($15$ corruptions of $5$ severity levels) associated common corruptions $\{\hat{\bm x}_n^c\}_{c=1}^C$, and generate for each of the clean images another $T=100$ transformed samples $\{\tilde{\bm x}^t_n\}_{t=1}^T$ using each augmentation scheme. Moreover, for each corruption $\hat{\bm x}_n^c$ we find its closest neighbor $\tilde{\bm x}_n^t$ from the set of generated samples using the cosine distance in the embedding space.
% \footnote{Examples of nearest neighbors can be found in~\cref{app:simclr_NNs}}. 
Our overall measure of fitness is
\begin{equation}
  \cfrac{1}{NC} \sum_{n=1}^N\sum_{c=1}^C\min_{t}\left\{1-\left(\cfrac{\phi(\hat{\bm x}^c_n)^\top\phi(\tilde{\bm x}^t_n)}{\|\phi(\hat{\bm x}^c_n)\|_2~\|\phi(\tilde{\bm x}^t_n)\|_2}\right)\right\}.
\end{equation}

\begin{table}[t]
% \resizebox{\columnwidth}{!}
% {%
    \centering
    \footnotesize
    \caption{Minimum cosine distances in the ResNet-50 SimCLRv2 embedding space between $100$ augmented samples from $1000$ ImageNet images, and their corresponding common corruptions.}
    \aboverulesep=0ex
    \belowrulesep=0ex
    \begin{tabular}{lcc}
        \MyToprule{1-3}
        \rule{0pt}{1.1EM}
        \multirow{2}{*}{Method} & \multicolumn{2}{c}{Min. cosine distance \footnotesize{($\times10^{-3}$)}} \\
        \MyMidrule{2-3}
        & Avg. ($\downarrow$) & Median  ($\downarrow$) \\
        \midrule
        None (clean) & 25.38 & 6.44 \\
        \midrule
        % \rule{0pt}{1.1EM}
        AugMix (w/o mix) & 20.57 & 3.56 \\
        PRIME (w/o mix) & \textbf{10.61} & \textbf{1.88} \\
        \midrule
        % \rule{0pt}{1.1EM}
        AugMix & 17.48 & 2.61 \\
        PRIME & \textbf{~~7.71} & \textbf{1.61} \\
        % \midrule
        % \rule{0pt}{1.1EM}
        % DA\texttt{+}AugMix & 15.44 & 1.93 & \multicolumn{1}{r}{6.93} & 4.26\\
        % DA\texttt{+}PRIME & \multicolumn{1}{r}{\textbf{7.69}} & \textbf{1.21} & \multicolumn{1}{r}{\textbf{4.14}} & \textbf{3.09}\\
        \bottomrule
    \end{tabular}
% }
    \vspace*{-2mm}
    \label{tab:simclr_distances}
\end{table}
\Cref{tab:simclr_distances} shows the values of this measure applied to AugMix and PRIME, with and without mixing. For reference, we also report the values of the clean (no transform) images $\{\bm x_n\}_{n=1}^N$. More percentile scores can be found in~\cref{app:simclr_cosine_stats}. Clearly, mixing helps reduce the distance between the common corruptions and the augmented samples from both methods. We also observe that PRIME, even with only $100$ augmentations per image -- in the order of the number of training epochs --  can generate samples that are twice as close to the common corruptions as AugMix. In fact, the feature similarity between training augmentations and test corruptions was also studied in~\cite{cbar2021}, with an attempt to justify the good performance of AugMix on C-10. Yet, we see that the fundamental transformations of AugMix are not enough to span a broad space guaranteeing high perceptual similarity to IN-C. The significant difference in terms of perceptual similarity in~\cref{tab:simclr_distances} between AugMix and PRIME may explain the superior performance of PRIME on IN-100-C and IN-C (cf.~\cref{tab:results-main})\footnote{A visualization of the augmented space using PCA can be found in~\cref{app:embedding-visual}.}.

\subsection{Robustness vs. accuracy trade-off}
\label{subsec:robustness-accuracy-tradeoff}

An important phenomenon observed in the literature of adversarial robustness is the so-called robustness-accuracy trade-off~\cite{Fawzi_Analysis_MachineLearning,tsiprasRobustnessMayBe2018,RaghunathanUnderstanding}, where technically adversarial training~\cite{madryDeepLearningModels2018} with smaller perturbations (typically smaller $\varepsilon$) results in models with higher standard but lower adversarial accuracy, and vice versa. In this sense, we want to understand if the strength of the image transformations introduced through data augmentations in PRIME can also cause such phenomenon in the context of robustness to common corruptions. As described in~\cref{sec:model_of_cc}, each of the transformations of PRIME has a strength parameter $\sigma$, which can be seen as the analogue of $\varepsilon$ in adversarial robustness. Hence, we can easily reduce or increase the strength of the transformations by setting $\hat{\sigma} = \alpha\sigma$, where $\alpha\in\R^+$. Then, by training a network for different values of $\alpha$ we can monitor its accuracy on the clean and the corrupted datasets.

We train a ResNet-18 on C-10 and IN-100 using the setup of~\cref{subsec:training_setup}. For reducing complexity, we do not use the JSD loss and train for $30$ epochs. This sub-optimal setting could cause some performance drop compared to the results of~\cref{tab:results-main}, but we expect the overall trends in terms of accuracy and robustness to be preserved. Regarding the scaling of the parameters' strength, for C-10 we set $\alpha\in[10^{-3}, 10^2]$ and sample $100$ values spaced evenly on a log-scale, while for IN-100 we set $\alpha\in[10^{-2}, 10^2]$ and we sample $20$ values.

\begin{figure}[t]
\centering
    \begin{minipage}[c]{0.40\linewidth}
    \centering
    \includegraphics[width=\linewidth]{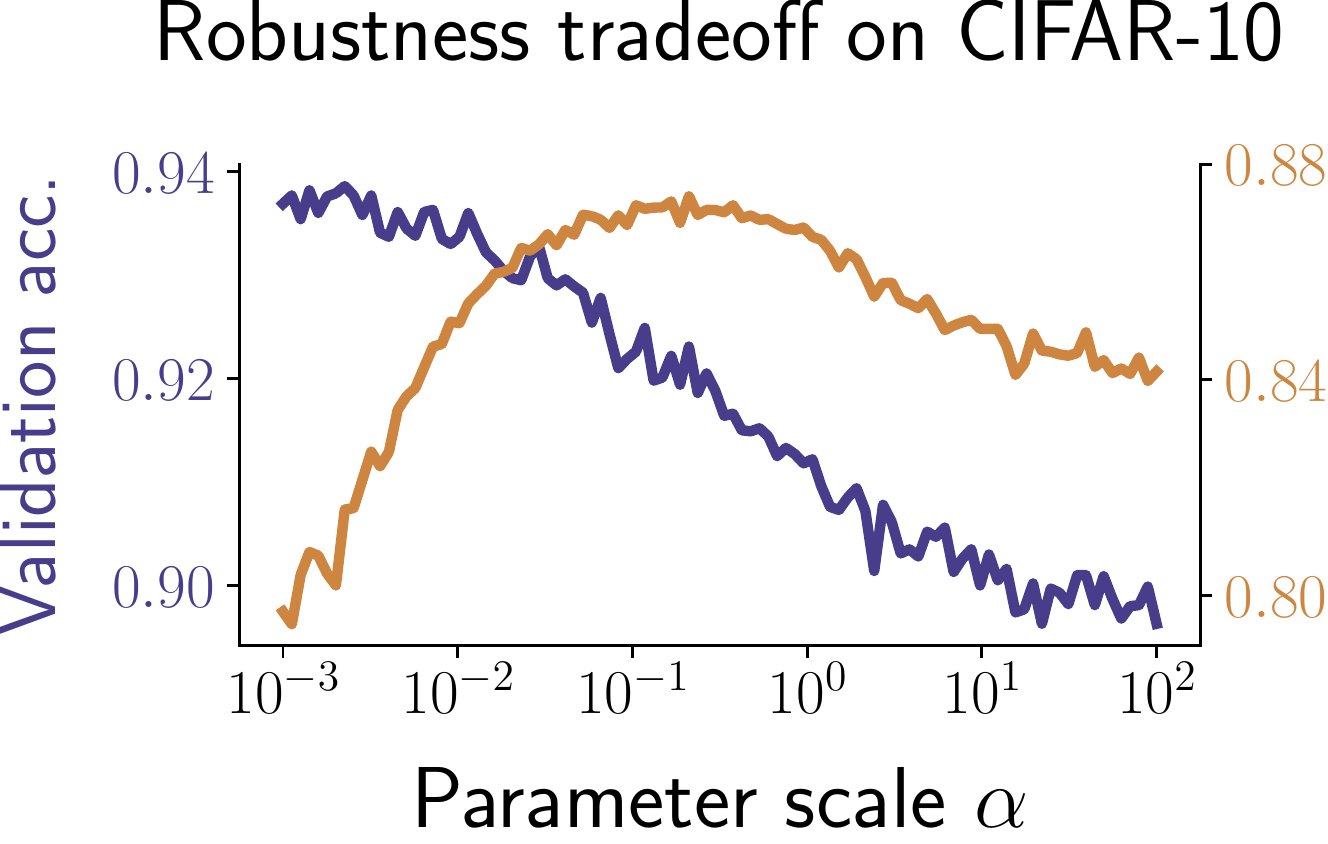} 
    % \vspace{1ex}
    \end{minipage}\;
    \begin{minipage}[c]{0.392\linewidth}
    \centering
    \includegraphics[width=\linewidth]{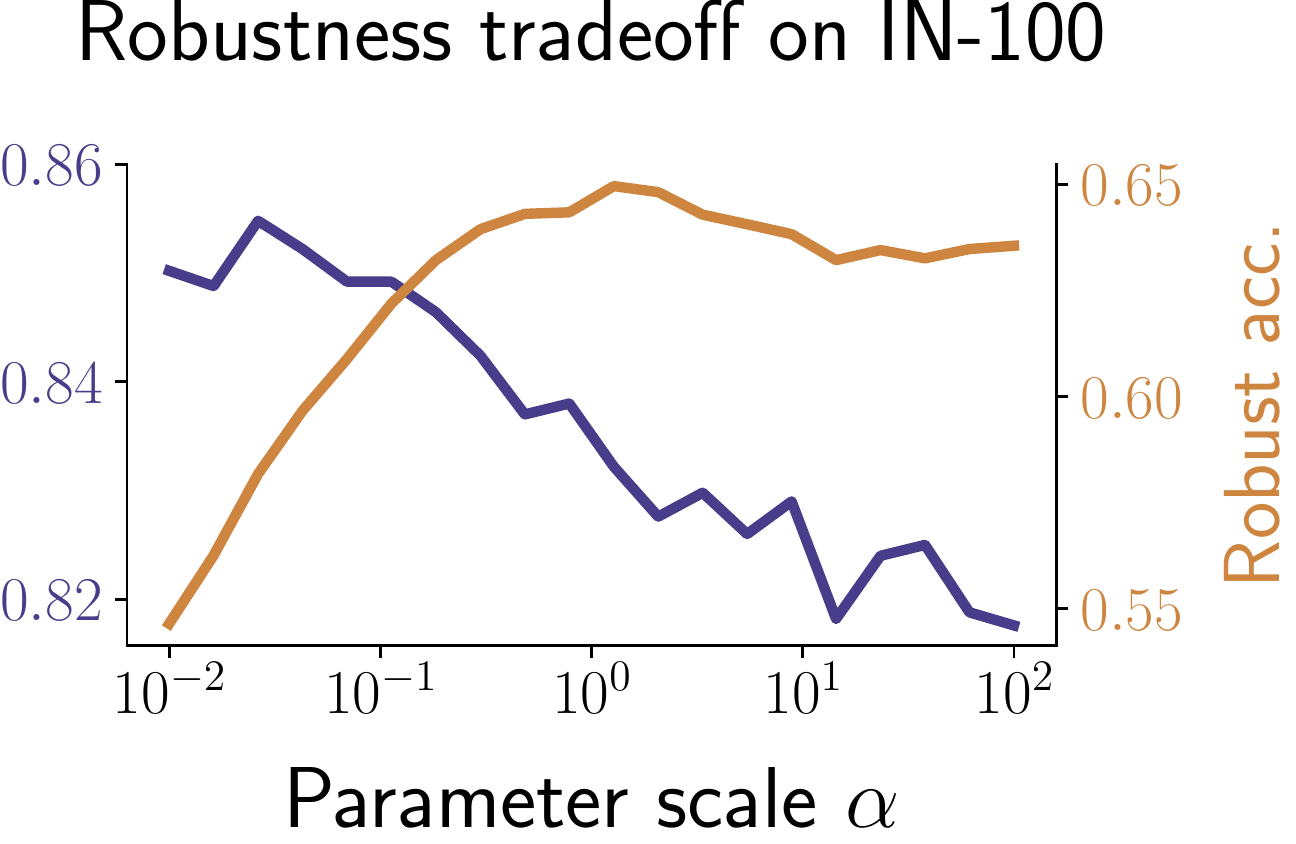}
    \end{minipage} 
    \caption{Robustness vs. accuracy of a ResNet-18 (w/o JSD) on CIFAR-10 (left) and ImageNet-100 (right), when trained multiple times with PRIME. On each training instance, the transformation strength is scaled by $\alpha$. Note the different scale in axes.}
    \vspace*{-2mm}
    \label{fig:rob_acc}
\end{figure}

The results are presented in~\cref{fig:rob_acc}. For both C-10 and IN-100, it seems that there is a sweet spot for the scale around $\alpha=0.2$ and $\alpha=1$ respectively, where the accuracy on common corruptions reaches its maximum. For $\alpha$ smaller than these values, we observe a clear trade-off between validation and robust accuracy. While the robustness to common corruptions increases, the validation accuracy decays. However, for $\alpha$ greater than the sweet-spot values, we observe that the trade-off ceases to exist since both the validation and robust accuracy present similar behaviour (slight decay). In fact, these observations indicate that robust and validation accuracies are not always positively correlated and that one might have to slightly sacrifice validation accuracy in order to achieve robustness.

\subsection{Sample complexity}
\label{subsec:sample-complexity}

Finally, we investigate the necessity of performing augmentation during training (on-line augmentation), compared to statically augmenting the dataset before training (off-line augmentation). On the one hand, on-line augmentation is useful when the dataset is huge and storing augmented versions requires a lot of memory. Besides, there are cases where offline augmentation is not feasible as it relies on pre-trained or generative models which are unavailable in certain scenarios, e.g., DeepAugment~\cite{deepaugment2021} or AdA~\cite{calian2021} cannot be applied on C-100. On the other hand, off-line augmentation may be necessary to avoid the computational cost of generating augmentations during training.

To this end, for each of the C-10 and IN-100 training sets, we augment them off-line with $k=1,2,\dots,10$ i.i.d. PRIME transformed versions. Afterwards, for different values of $k$, we train a ResNet-18 on the corresponding augmented dataset and report the accuracy on the validation set and the common corruptions. For the training setup, we follow the settings of~\cref{subsec:training_setup}, but without JSD loss. Also, since we increase the size of the training set by $(k+1)$, we also divide the number of training epochs by the same factor, in order to keep the same overall number of gradient updates.

\begin{figure}[t]
\centering
    \begin{minipage}[c]{0.37\columnwidth}
    \centering
    \includegraphics[width=\columnwidth]{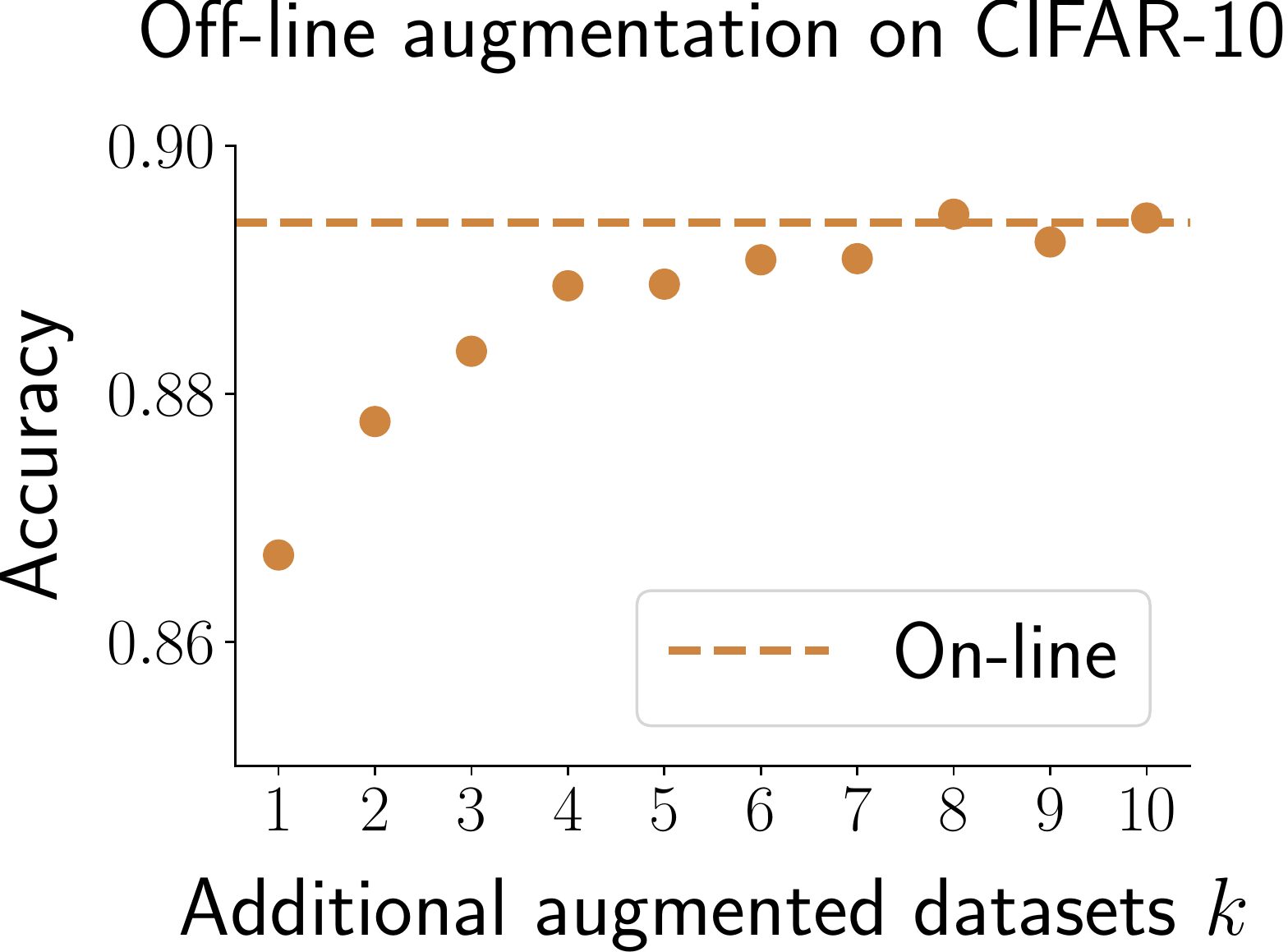}
    \label{fig:acc_vs_aug_cifar} 
    % \vspace{1ex}
    \end{minipage}
    \begin{minipage}[c]{0.355\columnwidth}
    \centering
    \includegraphics[width=\columnwidth]{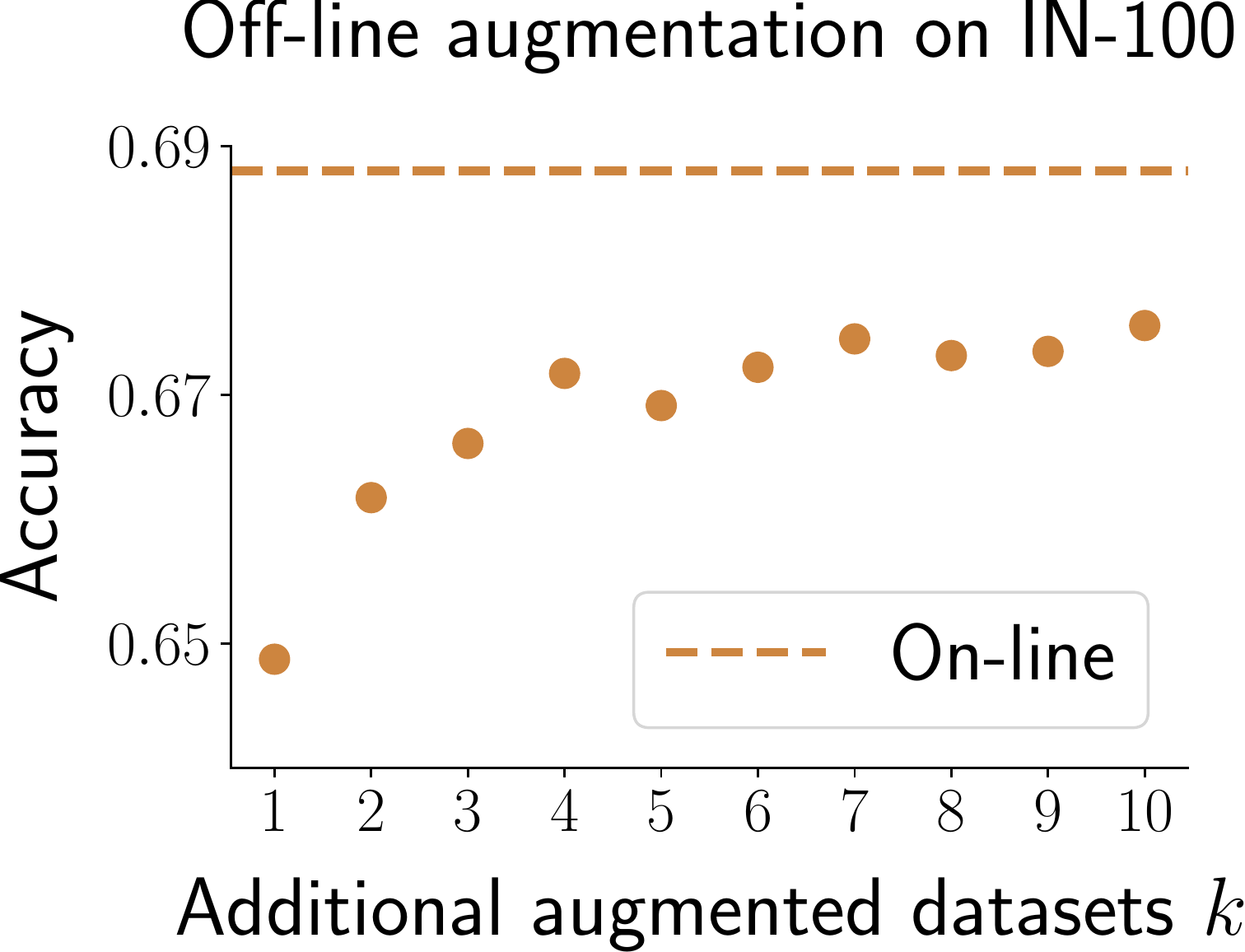}
    \label{fig:acc_vs_aug_imagenet} 
    % \vspace{1ex}
    \end{minipage}\hfill
    \caption{Accuracy of a ResNet-18 (w/o JSD) on CIFAR-10 (left) and ImageNet-100 (right) when augmenting the training sets with additional PRIME counterparts off-line. Dashed lines represent the accuracy achieved by training under the same setup, but generating the transformed samples during training (on-line augmentation). Validation accuracy is omitted because it is rather constant: around $93.4\%$ for CIFAR-10 and around $87\%$ for ImageNet-100.}
    \vspace*{-2mm}
    \label{fig:acc_vs_aug}
\end{figure}

The performance on common corruptions is presented in~\cref{fig:acc_vs_aug}. The first thing to notice is that, even for $k=1$, the obtained robustness to common corruptions is already quite good. In fact, for IN-100 the accuracy ($65\%$) is already better than AugMix ($60.7\%$ with JSD loss cf.~\cref{tab:results-main}). Regarding C-10, we observe that for $k=4$ the actual difference with respect to the on-line augmentation is almost negligible ($88.8\%$ vs. $89.3\%$), especially considering the overhead of transforming the data at every epoch. Technically, this means that augmenting C-10 with $4$ PRIME counterparts is enough for achieving good robustness to common corruptions. 
Finally, we also see in~\cref{fig:acc_vs_aug} that the corruption accuracy on IN-100 presents a very slow improvement after $k=4$. Comparing the accuracy at this point ($67.2\%$) to the one obtained with on-line augmentation and without JSD ($68.8\%$ cf. \cref{tab:orthogonality-in100}) we observe a gap of $1.6\%$. Hence, given the cost of on-line augmentation on such large scale datasets, simply augmenting the training with $4$ extra PRIME samples presents a good compromise for achieving competitive robustness. Nevertheless, the increase of $1.6\%$ introduced by on-line augmentation is rather significant, hinting that generating transformed samples during training might be necessary for maximizing performance. In this regard, the lower computational complexity of PRIME allows it to easily achieve this $+1.6\%$ gain through on-line augmentation, since it only requires $1.27\times$ additional training time compared to standard training, and only $1.06\times$ compared to AugMix, but with much better performance. This can be a significant advantage with respect to complex methods, like DeepAugment, that cannot be even applied on-line (require heavy pretraining).

%%%%%%%%%%%%%%%%%%%%%%%%%%%%%%%%%%%%%%
\section{Related work}
%%%%%%%%%%%%%%%%%%%%%%%%%%%%%%%%%%%%%%

\smallskip\noindent\textbf{Common corruptions} Towards evaluating the robustness of deep neural networks (DNNs) to natural distribution shifts, the authors in~\cite{corruptions2019} proposed common corruptions benchmarks (CIFAR-10-C and ImageNet-C) constituting 15 realistic image distortions. Later studies~\cite{deepaugment2021} considered the example of blurring and demonstrated that performance improvements on these common corruptions do generalize to real-world images, which supports the use of common corruptions benchmarks. Recent work~\cite{cbar2021} showed that current augmentation techniques undergo a performance degradation when evaluated on corruptions that are perceptually dissimilar from those in ImageNet-C. In addition to common corruptions, current literature studies other benchmarks e.g., adversarially filtered data~\cite{imageneta2021}, artistic renditions~\cite{deepaugment2021} and in-domain datasets~\cite{imagenetv22019}. In \cref{app:other_datasets}, we show that PRIME also improves robustness on these benchmarks.
% \pf{and? so what? what is the main message of this first paragraph?}

\smallskip\noindent\noindent\textbf{Improving corruption robustness} Data augmentation has been the central pillar for improving the generalization of DNNs~\cite{cutout2017,mixup2018,autoaugment2019,cutmix2019,patchgaussian2019}. A notable augmentation scheme for endowing corruption robustness is AugMix~\cite{augmix2020}, which employs a careful combination of stochastic augmentation operations and mixing. AugMix attains significant gains on CIFAR-10-C, but it does not perform as well on larger benchmarks like ImageNet-C. DeepAugment (DA)~\cite{deepaugment2021} addresses this issue and diversifies the space of augmentations by introducing distorted images computed by perturbing the weights of image-to-image networks. DA, combined with AugMix, achieves the current state-of-the-art on ImageNet-C. Other schemes include: (i) worst-case noise training~\cite{ant2020} or data augmentation through Fourier-based operations~\cite{Sun2021Certified}, (ii) inducing shape bias through stylized images~\cite{styleimagenet2018}, (iii) adversarial counterparts of DeepAugment~\cite{calian2021} and AugMix~\cite{augmax2021}, (iv) pre-training and/or adversarial training~\cite{yi2021,kireev2021}, (v) constraining the total variation of convolutional layers~\cite{tvmin2021} or compressing the model~\cite{Diffenderfer2021Winning} and (vi) learning the image information in the phase rather than amplitude~\cite{chenAmplitute2021} 
Besides, Vision Transformers~\cite{vits} have been shown to be more robust to common corruptions than standard CNNs~\cite{Bhojanapalli2021,morrison2021} when trained on big data. It would thus be interesting to study the effect of extra data alongside PRIME in future works. Finally, unsupervised domain adaptation~\cite{bnadapt_so2021,bnadapt_bethge2021} using a few corrupted samples has also been shown to provide a considerable boost in corruption robustness. Nonetheless, domain adaptation is orthogonal to this work as it requires knowledge of the target distribution.
% \pf{what is the main message of this second paragraph? Where do things stand? what are the limitations that we propsoe to overcome?}

%%%%%%%%%%%%%%%%%%%%%%%%%%%%%%%%%%%%%%
\section{Concluding remarks}
%%%%%%%%%%%%%%%%%%%%%%%%%%%%%%%%%%%%%%

We took a systematic approach to understand the notion of common corruptions and formulated a universal model that encompasses a wide variety of semantic-preserving image transformations. We then proposed a novel data augmentation scheme called \textit{PRIME}, which instantiates our model of corruptions, to confer robustness against common corruptions. From a practical perspective, our method is principled yet efficient and can be conveniently incorporated into existing training procedures. Moreover, it yields a strong baseline on existing corruption benchmarks outperforming current standalone methods. 
Additionally, our thorough ablations demonstrate that diversity among basic augmentations (primitives) -- which AugMix and other approaches lack -- is essential, and that mixing plays a crucial role in the success of both prior methods and PRIME.
In general, while complicated methods like DeepAugment perform well, it is difficult to understand, ablate and apply these online. Instead, we show that a simple model-based stance with a few guiding principles can be used to build a very effective augmentation scheme that can be easily understood, ablated and tuned.
We believe that our insights and PRIME pave the way for building robust models in real-life scenarios. PRIME, for instance, provides a ready-to-use recipe for data-scarce domains such as medical imaging.

%%%%%%%%%%%%%%%%%%%%%%%%%%%%%%%%%%%%%%
\section*{Acknowledgments}
%%%%%%%%%%%%%%%%%%%%%%%%%%%%%%%%%%%%%%

We thank Alessandro Favero for the fruitful discussions and feedback. This work has been partially supported by the CHIST-ERA program under Swiss NSF Grant 20CH21\_180444, and partially by Google via a Postdoctoral Fellowship and a GCP Research Credit Award.

\clearpage
\bibliographystyle{splncs04}
\bibliography{egbib}

\begin{thebibliography}{10}
\providecommand{\url}[1]{\texttt{#1}}
\providecommand{\urlprefix}{URL }
\providecommand{\doi}[1]{https://doi.org/#1}

\bibitem{beale}
Beale, P.: Statistical Mechanics. Elsevier (1996)

\bibitem{bnadapt_so2021}
Benz, P., Zhang, C., Karjauv, A., Kweon, I.S.: Revisiting batch normalization
  for improving corruption robustness. In: Proceedings of the IEEE/CVF Winter
  Conference on Applications of Computer Vision (2021)

\bibitem{Bhojanapalli2021}
Bhojanapalli, S., Chakrabarti, A., Glasner, D., Li, D., Unterthiner, T., Veit,
  A.: Understanding robustness of {Transformers} for image classification. In:
  Proceedings of the IEEE/CVF International Conference on Computer Vision
  (ICCV) (2021)

\bibitem{near-ring}
Binder, F., Aichinger, E., Ecker, J., N\"{o}bauer, C., Mayr, P.: Algorithms for
  near-rings of non-linear transformations. In: Proceedings of the
  International Symposium on Symbolic and Algebraic Computation. Association
  for Computing Machinery (2000)

\bibitem{calian2021}
Calian, D.A., Stimberg, F., Wiles, O., Rebuffi, S.A., Gyorgy, A., Mann, T.,
  Gowal, S.: Defending against image corruptions through adversarial
  augmentations. arXiv preprint arXiv:2104.01086  (2021)

\bibitem{chenAmplitute2021}
Chen, G., Peng, P., Ma, L., Li, J., Du, L., Tian, Y.: Amplitude-phase
  recombination: {Rethinking} robustness of convolutional neural networks in
  frequency domain. In: Proceedings of the IEEE/CVF International Conference on
  Computer Vision (ICCV) (2021)

\bibitem{simclr}
Chen, T., Kornblith, S., Swersky, K., Norouzi, M., Hinton, G.E.: Big
  self-supervised models are strong semi-supervised learners. In: Advances in
  Neural Information Processing Systems (2020)

\bibitem{cover_info}
Cover, T.M., Thomas, J.A.: Elements of Information Theory. Wiley-Interscience
  (2006)

\bibitem{robustbench2021}
Croce, F., Andriushchenko, M., Sehwag, V., Debenedetti, E., Flammarion, N.,
  Chiang, M., Mittal, P., Hein, M.: Robustbench: a standardized adversarial
  robustness benchmark. In: Thirty-fifth Conference on Neural Information
  Processing Systems Datasets and Benchmarks Track (2021)

\bibitem{autoaugment2019}
Cubuk, E.D., Zoph, B., Mané, D., Vasudevan, V., Le, Q.V.: Autoaugment:
  Learning augmentation strategies from data. In: 2019 IEEE/CVF Conference on
  Computer Vision and Pattern Recognition (2019)

\bibitem{imagenet2009}
Deng, J., Dong, W., Socher, R., Li, L.J., Li, K., Fei-Fei, L.: Imagenet: A
  large-scale hierarchical image database. In: 2009 IEEE Conference on Computer
  Vision and Pattern Recognition (2009)

\bibitem{cutout2017}
DeVries, T., Taylor, G.W.: Improved regularization of convolutional neural
  networks with cutout. arXiv preprint arXiv:1708.04552  (2017)

\bibitem{Diffenderfer2021Winning}
Diffenderfer, J., Bartoldson, B.R., Chaganti, S., Zhang, J., Kailkhura, B.: A
  winning hand: Compressing deep networks can improve out-of-distribution
  robustness. In: Advances in Neural Information Processing Systems (Dec 2021)

\bibitem{dodgekaram2016}
Dodge, S., Karam, L.: Understanding how image quality affects deep neural
  networks. In: 2016 Eighth International Conference on Quality of Multimedia
  Experience (QoMEX) (2016)

\bibitem{vits}
Dosovitskiy, A., Beyer, L., Kolesnikov, A., Weissenborn, D., Zhai, X.,
  Unterthiner, T., Dehghani, M., Minderer, M., Heigold, G., Gelly, S.,
  Uszkoreit, J., Houlsby, N.: An image is worth 16x16 words: {Transformers} for
  image recognition at scale. In: International Conference on Learning
  Representations (2021)

\bibitem{Fawzi_Analysis_MachineLearning}
Fawzi, A., Fawzi, O., Frossard, P.: Analysis of classifiers' robustness to
  adversarial perturbations. Machine Learning  \textbf{107}(3),  481--508
  (2018)

\bibitem{styleimagenet2018}
Geirhos, R., Rubisch, P., Michaelis, C., Bethge, M., Wichmann, F.A., Brendel,
  W.: Imagenet-trained {CNN}s are biased towards texture; increasing shape bias
  improves accuracy and robustness. In: International Conference on Learning
  Representations (2019)

\bibitem{humans2018}
Geirhos, R., Temme, C.R.M., Rauber, J., Sch\"{u}tt, H.H., Bethge, M., Wichmann,
  F.A.: Generalisation in humans and deep neural networks. In: Advances in
  Neural Information Processing Systems (2018)

\bibitem{resnet2016}
{He}, K., {Zhang}, X., {Ren}, S., {Sun}, J.: Deep residual learning for image
  recognition. In: 2016 IEEE Conference on Computer Vision and Pattern
  Recognition (2016)

\bibitem{deepaugment2021}
Hendrycks, D., Basart, S., Mu, N., Kadavath, S., Wang, F., Dorundo, E., Desai,
  R., Zhu, T., Parajuli, S., Guo, M., Song, D., Steinhardt, J., Gilmer, J.: The
  many faces of robustness: A critical analysis of out-of-distribution
  generalization. In: IEEE Conference on Computer Vision and Pattern
  Recognition (2021)

\bibitem{corruptions2019}
Hendrycks, D., Dietterich, T.: Benchmarking neural network robustness to common
  corruptions and perturbations. In: International Conference on Learning
  Representations (2019)

\bibitem{augmix2020}
Hendrycks*, D., Mu*, N., Cubuk, E.D., Zoph, B., Gilmer, J., Lakshminarayanan,
  B.: Augmix: A simple method to improve robustness and uncertainty under data
  shift. In: International Conference on Learning Representations (2020)

\bibitem{imageneta2021}
Hendrycks, D., Zhao, K., Basart, S., Steinhardt, J., Song, D.: Natural
  adversarial examples. In: Proceedings of the IEEE/CVF Conference on Computer
  Vision and Pattern Recognition (2021)

\bibitem{kireev2021}
Kireev, K., Andriushchenko, M., Flammarion, N.: On the effectiveness of
  adversarial training against common corruptions. arXiv preprint
  arXiv:2103.02325  (2021)

\bibitem{cifar102009}
Krizhevsky, A.: Learning multiple layers of features from tiny images (2009)

\bibitem{patchgaussian2019}
Lopes, R.G., Yin, D., Poole, B., Gilmer, J., Cubuk, E.D.: Improving robustness
  without sacrificing accuracy with patch gaussian augmentation. arXiv preprint
  arXiv:1906.02611  (2019)

\bibitem{madryDeepLearningModels2018}
Madry, A., Makelov, A., Schmidt, L., Tsipras, D., Vladu, A.: Towards deep
  learning models resistant to adversarial attacks. In: International
  {{Conference}} on {{Learning Representations}} (Apr 2018)

\bibitem{OODInfoBound}
Masiha, M.S., Gohari, A., Yassaee, M.H., Aref, M.R.: Learning under
  distribution mismatch and model misspecification. In: {IEEE} International
  Symposium on Information Theory, {(ISIT)} (2021)

\bibitem{cbar2021}
Mintun, E., Kirillov, A., Xie, S.: On interaction between augmentations and
  corruptions in natural corruption robustness. arXiv preprint arXiv:2102.11273
   (2021)

\bibitem{moayeri_sample_efficient_iccv}
Moayeri, M., Feizi, S.: Sample efficient detection and classification of
  adversarial attacks via self-supervised embeddings. In: Proceedings of the
  IEEE/CVF International Conference on Computer Vision (ICCV) (2021)

\bibitem{morrison2021}
Morrison, K., Gilby, B., Lipchak, C., Mattioli, A., Kovashka, A.: Exploring
  corruption robustness: {Inductive} biases in vision transformers and
  mlp-mixers. arXiv preprint arXiv:2106.13122  (2021)

\bibitem{nesterov1983}
Nesterov, Y.E.: A method for solving the convex programming problem with
  convergence rate ${O}(1/k^2)$. Dokl. Akad. Nauk SSSR  (1983)

\bibitem{torch2019}
Paszke, A., Gross, S., Massa, F., Lerer, A., Bradbury, J., Chanan, G., Killeen,
  T., Lin, Z., Gimelshein, N., Antiga, L., Desmaison, A., Kopf, A., Yang, E.,
  DeVito, Z., Raison, M., Tejani, A., Chilamkurthy, S., Steiner, B., Fang, L.,
  Bai, J., Chintala, S.: Pytorch: An imperative style, high-performance deep
  learning library. In: Advances in Neural Information Processing Systems
  (2019)

\bibitem{diffeo}
Petrini, L., Favero, A., Geiger, M., Wyart, M.: Relative stability toward
  diffeomorphisms indicates performance in deep nets. In: Advances in Neural
  Information Processing Systems (2021)

\bibitem{RaghunathanUnderstanding}
Raghunathan, A., Xie, S.M., Yang, F., Duchi, J., Liang, P.: Understanding and
  mitigating the tradeoff between robustness and accuracy. In: Proceedings of
  the 37th International Conference on Machine Learning (Jul 2020)

\bibitem{imagenetv22019}
Recht, B., Roelofs, R., Schmidt, L., Shankar, V.: Do {I}mage{N}et classifiers
  generalize to {I}mage{N}et? In: Proceedings of the 36th International
  Conference on Machine Learning (2019)

\bibitem{ant2020}
Rusak, E., Schott, L., Zimmermann, R.S., Bitterwolf, J., Bringmann, O., Bethge,
  M., Brendel, W.: A simple way to make neural networks robust against diverse
  image corruptions. In: Computer Vision -- ECCV 2020 (2020)

\bibitem{tvmin2021}
Saikia, T., Schmid, C., Brox, T.: Improving robustness against common
  corruptions with frequency biased models. In: Proceedings of the IEEE/CVF
  International Conference on Computer Vision (ICCV) (2021)

\bibitem{bnadapt_bethge2021}
Schneider, S., Rusak, E., Eck, L., Bringmann, O., Brendel, W., Bethge, M.:
  Improving robustness against common corruptions by covariate shift
  adaptation. In: Advances in Neural Information Processing Systems (2020)

\bibitem{cyclic2018}
Smith, L.N., Topin, N.: Super-convergence: Very fast training of residual
  networks using large learning rates. arXiv preprint arXiv:1708.07120  (2018)

\bibitem{Sun2021Certified}
Sun, J., Mehra, A., Kailkhura, B., Chen, P.Y., Hendrycks, D., Hamm, J., Mao,
  Z.M.: Certified adversarial defenses meet out-of-distribution corruptions:
  {Benchmarking} robustness and simple baselines. arXiv preprint
  arXiv:arXiv:2112.00659  (2021)

\bibitem{taori2020}
Taori, R., Dave, A., Shankar, V., Carlini, N., Recht, B., Schmidt, L.:
  Measuring robustness to natural distribution shifts in image classification.
  In: Advances in Neural Information Processing Systems (2020)

\bibitem{tsiprasRobustnessMayBe2018}
Tsipras, D., Santurkar, S., Engstrom, L., Turner, A., Madry, A.: Robustness may
  be at odds with accuracy. In: International Conference on Learning
  Representations (May 2019)

\bibitem{augmax2021}
Wang, H., Xiao, C., Kossaifi, J., Yu, Z., Anandkumar, A., Wang, Z.: Augmax:
  Adversarial composition of random augmentations for robust training. In:
  Advances in Neural Information Processing Systems (2021)

\bibitem{xuInfoBound}
Xu, A., Raginsky, M.: Information-theoretic analysis of generalization
  capability of learning algorithms. In: Advances in Neural Information
  Processing Systems (2017)

\bibitem{yi2021}
Yi, M., Hou, L., Sun, J., Shang, L., Jiang, X., Liu, Q., Ma, Z.: Improved {OOD}
  generalization via adversarial training and pretraing. In: Proceedings of the
  86th International Conference on Machine Learning (2021)

\bibitem{cutmix2019}
Yun, S., Han, D., Chun, S., Oh, S.J., Yoo, Y., Choe, J.: Cutmix: Regularization
  strategy to train strong classifiers with localizable features. In: 2019
  IEEE/CVF International Conference on Computer Vision (2019)

\bibitem{mixup2018}
Zhang, H., Cisse, M., Dauphin, Y.N., Lopez-Paz, D.: mixup: Beyond empirical
  risk minimization. In: International Conference on Learning Representations
  (2018)

\bibitem{lpips}
Zhang, R., Isola, P., Efros, A.A., Shechtman, E., Wang, O.: The unreasonable
  effectiveness of deep features as a perceptual metric. In: 2018 IEEE/CVF
  Conference on Computer Vision and Pattern Recognition (2018)

\end{thebibliography}

\clearpage
\appendix

\section{Maximum entropy transformations}
\label{app:max-entropy}

To guarantee as much diversity as possible in our model of common corruptions, we follow the principle of maximum entropy to define our distributions of transformations~\cite{cover_info}. Note that using a set of augmentations that guarantees maximum entropy comes naturally when trying to optimize the sample complexity derived from certain information theoretic generalization bounds, both in the clean~\cite{xuInfoBound} and corrupted setting~\cite{OODInfoBound}. Specifically, the principle of maximum entropy postulates favoring those distributions that are as unbiased as possible given the set of constraints that defines a family of distributions. In our case, these constraints are given in the form of an expected strength, i.e., $\sigma^2$, desired smoothness, i.e., $K$, and/or some boundary conditions, e.g.,, the displacement field must be zero at the borders of an image.  

Let us make this formal. In particular, let $\mathcal{I}$ denote the space of all images $\bm x:\R^2\to\R^3$, and let $f:\mathcal{I}\to \mathcal{I}$ denote a random image transformation distributed according to the law $\mu$. Further, let us define a set of constraints $\mathcal{C}\subseteq \mathcal{F}$, which restrict the domain of applicability of $f$, i.e., $f\in\mathcal{C}$, and where $\mathcal{F}$ denotes the space of functions $\mathcal{I}\to \mathcal{I}$. The principle of maximum entropy postulates using the distribution $\mu$ which has maximum entropy given the constraints:
\begin{align}
    \underset{\mu}{\text{maximize}}\quad &  H(\mu)=\int_{\mathcal{F}}\, \mathrm{d}\mu(f) \log(\mu(f)) \label{eq:app-max_entropy}\\
    \text{subject to}\quad & f\in\mathcal C\quad \forall f\sim\mu, \nonumber
\end{align}
where $H(\mu)$ represents the entropy of the distribution $\mu$~\cite{cover_info}. In its general form, solving \cref{eq:app-max_entropy} for any set of constraints $\mathcal{C}$ is intractable. However, leveraging results from statistical physics, we will see that for our domains of interest, \cref{eq:app-max_entropy} has a simple solution. In what follows we derive those distributions for each of our family of transformations.

\subsection{Spectral domain}

As we introduced in \cref{sec:model_of_cc}, we propose to parameterize our family of spectral transformations using an FIR filter of size $K_\omega\times K_\omega$. That is, we are interested in finding a maximum entropy distribution over the space of spectral transformations with a finite spatial support. 

Nevertheless, on top of this smoothness constraint we are also interested in controlling the strength of the transformations.  We define the strength of a distribution of random spectral transformations applied to an image $\bm x$, as the expected $L_2$ norm of the difference between the clean and transformed images, i.e.,
\begin{equation}
  \mathbb{E}_{\omega} \|\bm x-\omega(\bm x)\|^2_2=\mathbb{E}_{\bm\omega'}\|\bm \omega' * \bm x\|_2^2,
\end{equation}
which using Young's convolution inequality is bounded as
\begin{equation}
  \mathbb{E}_{\bm\omega'}\|\bm \omega' * \bm x\|_2^2\leq \|\bm x\|_1^2 \;\mathbb{E}_{\bm \omega'}\|\bm\omega'\|_2^2.
\end{equation}
Indeed, we can see that the strength of a distribution of random smooth spectral transformations is governed by the expected norm of its filter. In the discrete domain, this can be simply computed as
\begin{equation}
  \mathbb{E}_{\bm \omega'}\|\bm\omega'\|_2^2=\sum_{i=1}^{K_\omega}\sum_{j=1}^{K_\omega}\mathbb{E}_{\bm\omega'}\bm {\omega'}^2_{i,j}.\label{eq:norm_omega}
\end{equation}

Considering this, we should then look for a maximum entropy distribution whose samples satisfy
\begin{equation}
    \mathcal{C}=\left\{\bm\omega'\in\R^{K_\omega\times K_\omega}\wedge \mathbb{E}_{\bm\omega'}\|\bm\omega'\|_2^2=K_\omega^2 \sigma^2_\omega \,|\, \omega\sim \mu_\omega\right\}.\label{eq:spectral_constraint}
\end{equation}

Now, note that this set is defined by an equality constraint involving a sum of $K_\omega^2$ quadratic random variables. In this sense, we know that the Equipartition Theorem~\cite{beale} applies and can be used to identify the distribution of maximum entropy. That is, the solution of \cref{eq:app-max_entropy} in the case that $\mathcal{C}$ is given by \cref{eq:spectral_constraint}, is equal to the distribution of FIR filters whose coefficients are iid with law $\mathcal{N}(0, \sigma^2_\omega)$.

\subsection{Spatial domain}
The distribution of diffeomorphisms of maximum entropy with a fixed norm was derived by Petrini \etal in \cite{diffeo}. The derivation is similar to the spectral domain, but with the additional constraint that the diffeomorphisms produce a null displacement at the borders of the image.

\subsection{Color domain}
We can follow a very similar route to derive the distribution of maximum entropy among all color transformations, where, specifically, we constraint the transformations to yield $\gamma(0)=0$ and $\gamma(1)=1$ on every channel independently. Doing so, the derivation of the maximum entropy distribution can follow the same steps as in \cite{diffeo}.

\section{PRIME implementation details}
\label{app:implementation_details}
In this section, we provide additional details regarding the implementation of PRIME described in \cref{sec:prime}. Since the parameters of the transformations are empirically selected, we first provide more visual examples for different values of smoothness $K$ and strength $\sigma$. Then, we give the exact values of the parameters we use in our experiments supported by additional visual examples and we also describe the parameters we use for the mixing procedure.

\subsection{Additional transformed examples}

\begin{figure}[!ht]
    \centering
    \includegraphics[width=0.75\columnwidth]{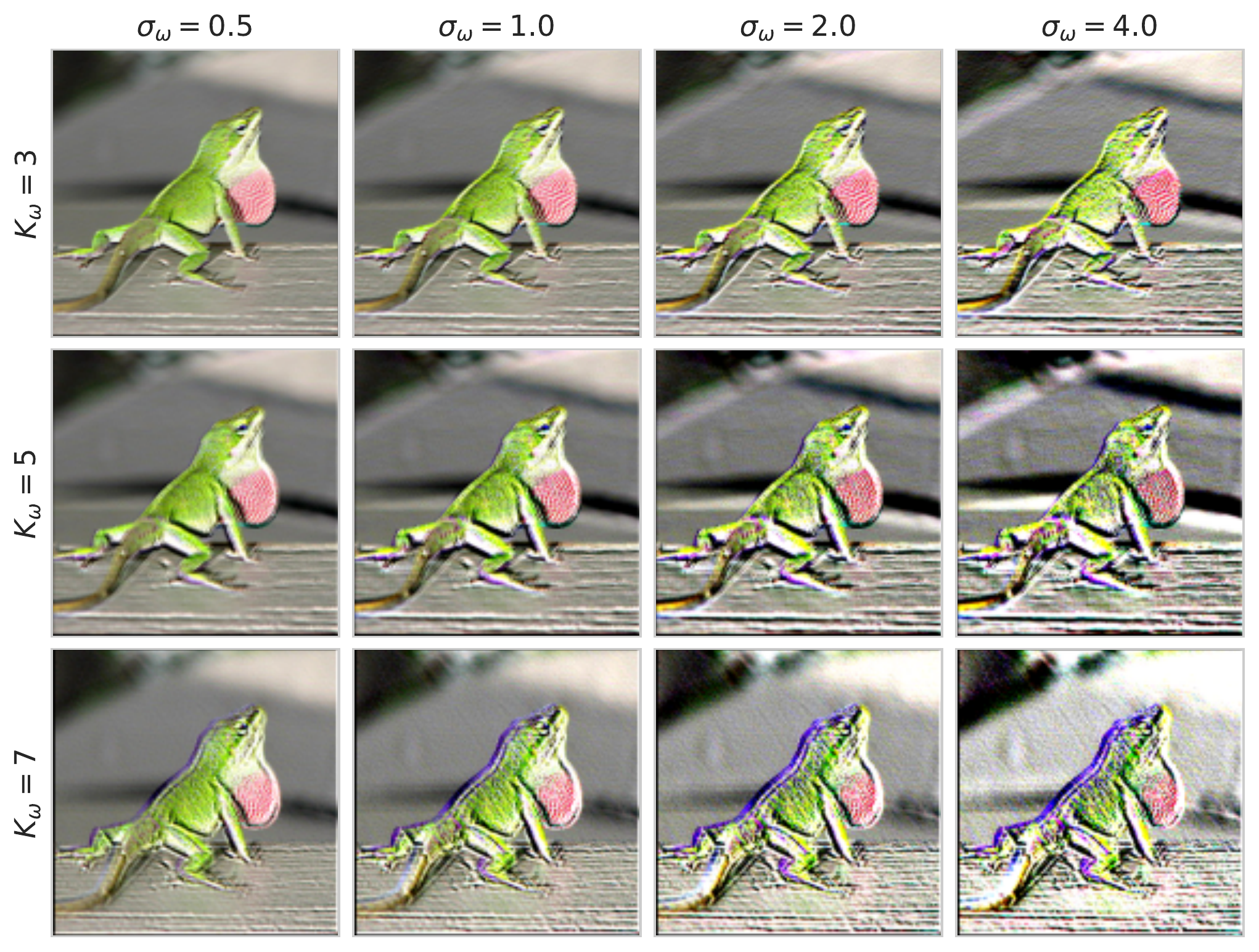}
    \caption{Example images (IN) generated with spectral transformations from our common corruptions model. In each row, we enlarge the transformation strength $\sigma_{\omega}$ from left to right. From top to bottom, we increase the spectral resolution of the filter $K_\omega$.}
    \label{fig:ex_spectral}
\end{figure}

\begin{figure}[!ht]
    \centering
    \includegraphics[width=0.75\columnwidth]{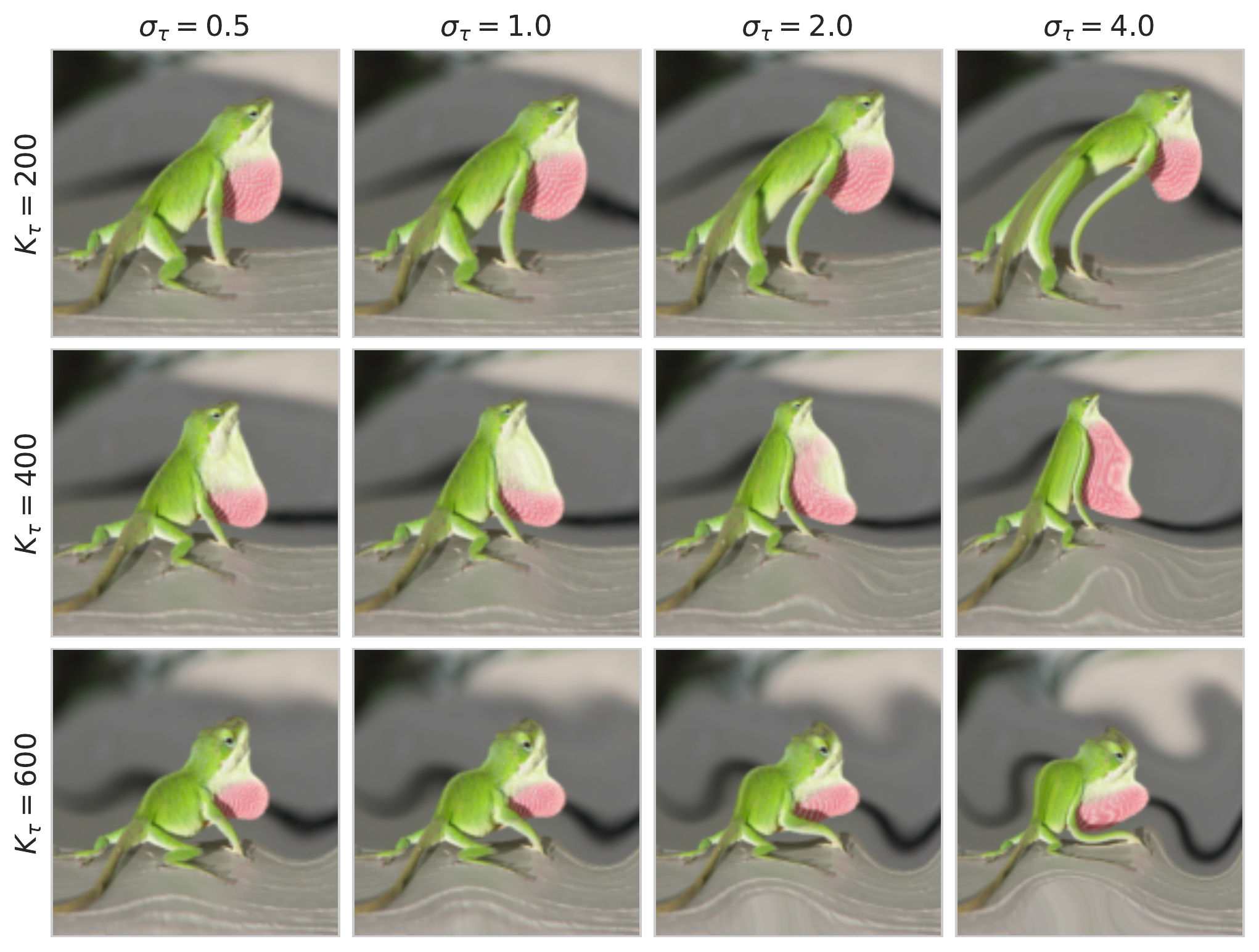}
    \caption{Example images (IN) generated with spatial transformations from our common corruptions model. In each row, we enlarge the transformation strength $\sigma_{\tau}$ from left to right. From top to bottom, we increase the cut frequency $K_\tau$.}
    \label{fig:ex_spatial}
\end{figure}

\begin{figure}[!ht]
    \centering
    \includegraphics[width=0.75\columnwidth]{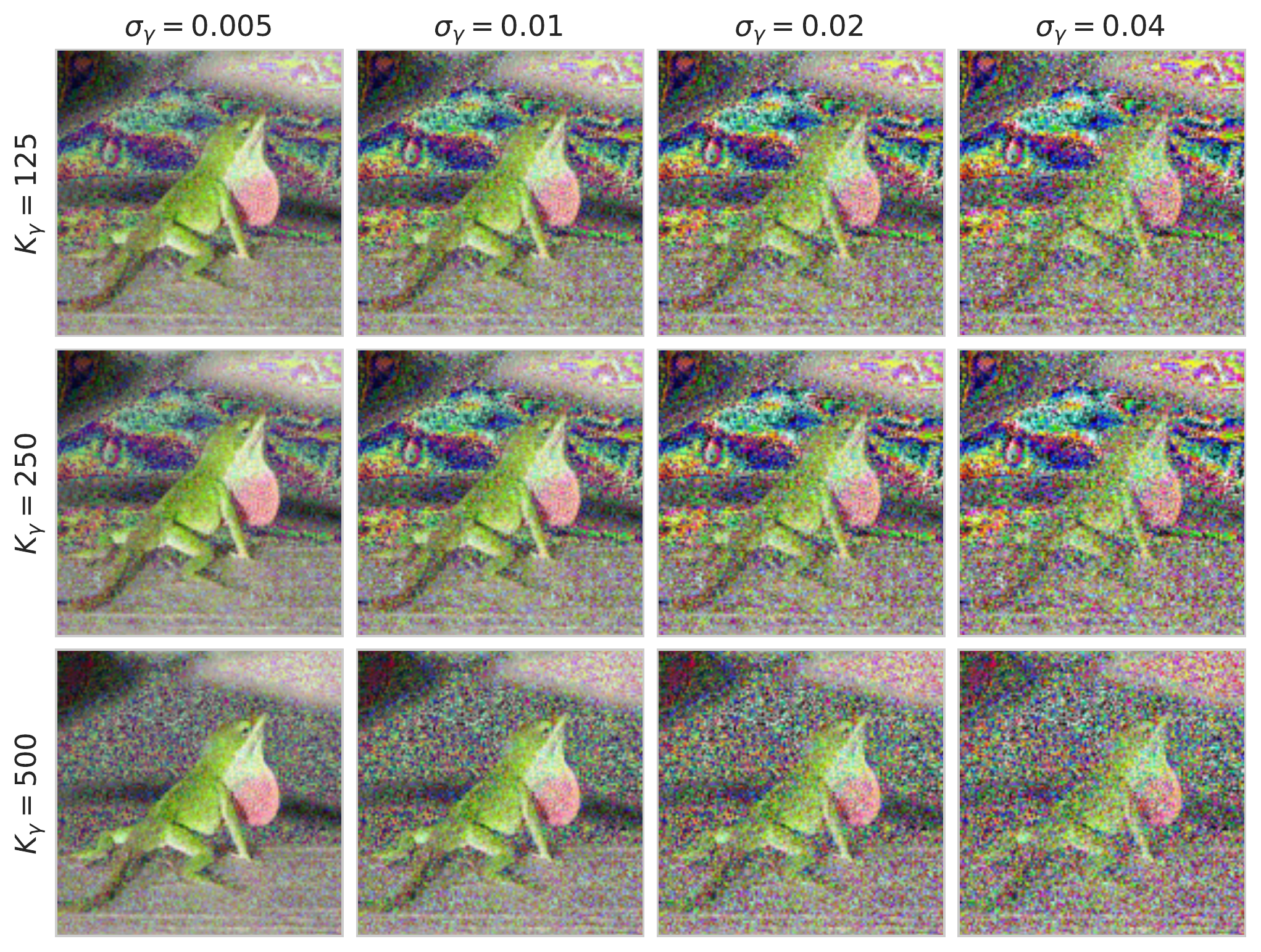}
    \caption{Example images (IN) generated with color transformations from our common corruptions model. In each row, we enlarge the transformation strength $\sigma_{\gamma}$ from left to right. From top to bottom, we increase $K_\gamma$.}
    \vspace{-0.8em}
\label{fig:ex_color}
\end{figure}

We provide additional visual examples for each of the primitives of PRIME illustrating the effect of the following two factors: (i) smoothness controlled by parameter $K$, and (ii) strength of the transformation $\sigma$ on the resulting transformed images created by the primitives. \cref{fig:ex_spectral,,fig:ex_spatial,,fig:ex_color} demonstrate the resulting spectrum of images created by applying spectral, spatial and color transformations while varying the parameters $K$ and $\sigma$. Notice how increasing the strength $\sigma$ of each transformation drifts the augmented image farther away from its clean counterpart, yet produces plausible images when appropriately controlled.

\clearpage
\newpage

\subsection{Transformation parameters}
We now provide the parameters of each transform that we selected and used in our experiments. In general, the values might vary for inputs of different dimensionality and resolution (i.e., CIFAR-10/100 vs ImageNet images).

\subsubsection{Spectral transform}
Regarding the spectral transform of \cref{eq:spectral_domain} we found out that, for the FIR filter $\bm \omega'$, a size of $K_\omega=3$ results into semantically preserving images for CIFAR-10/100 and ImageNet. For the latter, one can stretch the filter size to $5\times5$ or even $7\times7$, but then slight changes on the strength, $\sigma_\omega$, might destroy the image semantics. Eventually, given $K_\omega=3$, we observed that $\sigma_\omega=4$ is good enough for CIFAR-10/100 and ImageNet. 

\subsubsection{Spatial transform}
Concerning the spatial transform of \cref{eq:spatial_transform}, for the cut-off parameter $K_\tau$ we followed the value regimes proposed by Petrini \etal~\cite{diffeo} and set $K_\tau=100$ for CIFAR-10/100; $K_\tau=500$ for ImageNet. Furthermore, for a given $K_\tau$, Petrini \etal also compute the appropriate bounds for the transformation strength, $\sigma^2_{\tau_\text{min}} \leq \sigma^2_\tau \leq \sigma^2_{\tau_\text{max}}$, such that the resulting diffeomorphism remains bijective and the pixel displacement does not destroy the image. In fact, in their original implementation\footnote{The official implementation of Petrini \etal diffeomorphisms can be found at \url{https://github.com/pcsl-epfl/diffeomorphism}.}, Petrini \etal directly sample $\sigma_\tau\sim U(\sigma_{\tau_\text{min}},\sigma_{\tau_\text{max}})$ instead of explicitly setting the strength. In our implementation, we also follow the same approach. 

\subsubsection{Color transform}
Regarding the color transform of \cref{eq:color_transform} we found out that for CIFAR-10/100 a cut-off value of $K_\gamma=10$ and a strength of $\sigma_\gamma=0.01$ result into semantically preserving images for CIFAR-10/100; while for ImageNet, the corresponding values are $K_\gamma=500$ and $\sigma_\gamma=0.05$. As for the bandwidth (consecutive frequencies) $\Delta$ we observed that a value of $\Delta=20$ was memory sufficient for ImageNet, but for CIFAR-10/100, due to its lower dimensionality, we can afford all the frequencies to be used, e.g., $\Delta=K_\gamma$. 

\begin{figure}[t]
    \centering
    \includegraphics[width=0.55\columnwidth]{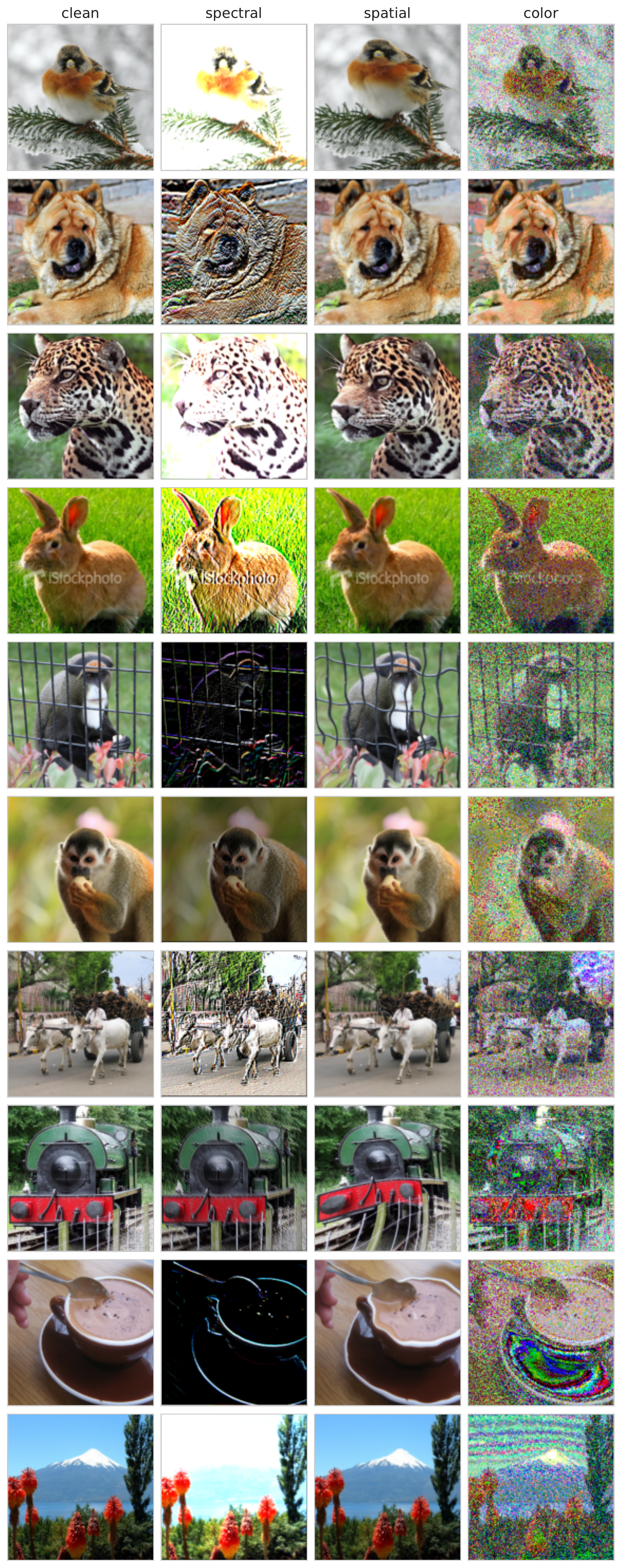}
    \caption{Example images (IN) generated with the transformations of our common corruptions model. Despite the perceptibility of the introduced distortion, the image semantics are preserved.}
    \vspace{-0.8em}
\label{fig:extra_ex_max_ent}
\end{figure}

\bigskip\medskip\noindent
Finally, as mentioned in \cref{sec:prime}, we randomly sample the strength of the transformations $\sigma$ from a uniform distribution of given minimum and maximum values. Regarding the maximum, we always set it to be the one we selected through visual inspection, while the minimum is set to $0$. \cref{fig:extra_ex_max_ent} displays additional augmented images created by applying each of the primitive transformations in our model using the aforementioned set of parameters on ImageNet. Our choice of parameters produces diverse image augmentations, while retaining the semantic content of the images.

\clearpage
\newpage

\subsection{Parameters for mixing procedure}
Regarding the mixing parameters of our experiments, we fix the total number of generated transformed images (width) to be $n=3$. As for the composition of the transformations (depth), we follow a stochastic approach such that, on every iteration $i\in\{1,\dots,n\}$, only $\hat{m}\in[1,m]$ compositions are performed, with $m=3$. In fact, in \cref{alg:prime} we do not explicitly select randomly a new $\hat{m}$ for every $i$ but we provide the identity operator $\operatorname{Id}$ instead. This guarantees that, in some cases, no transformation is performed.

\section{Detailed experimental setup}
\label{app:exp-setup}

We now provide all the experimental details for the performance evaluation of \cref{sec:experimental}. All models are implemented in PyTorch~\cite{torch2019} and are trained for $100$ epochs using a cyclic learning rate schedule~\cite{cyclic2018} with cosine annealing and a maximum learning rate of $0.2$ unless stated otherwise. For IN, we fine-tune a regularly pretrained network (provided in PyTorch) with a maximum learning rate of $0.01$ following Hendrycks \etal~\cite{deepaugment2021}. We use SGD optimizer with momentum factor $0.9$ and Nesterov momentum~\cite{nesterov1983}. On C-10 \& C-100, we set the batch size to $128$ and use a weight decay of $0.0005$. On IN-100 and IN, the batch size is $256$ and weight decay is $0.0001$. We employ ResNet-18~\cite{resnet2016} on C-10, C-100 and IN-100; and use ResNet-50 for IN. The augmentation hyperparameters for AugMix and DeepAugment are the same as in their original implementations. 
% The code to reproduce our experiments is publicly available at \url{https://github.com/amodas/PRIME-augmentations}.

\section{Additional mixing examples}
\label{app:mixing_examples}

\begin{figure}[!ht]
    \centering
    \footnotesize
    PRIME\\\vspace{3pt}
    \begin{subfigure}[t]{0.395\columnwidth}
        \centering
        \footnotesize
        \includegraphics[width=\linewidth]{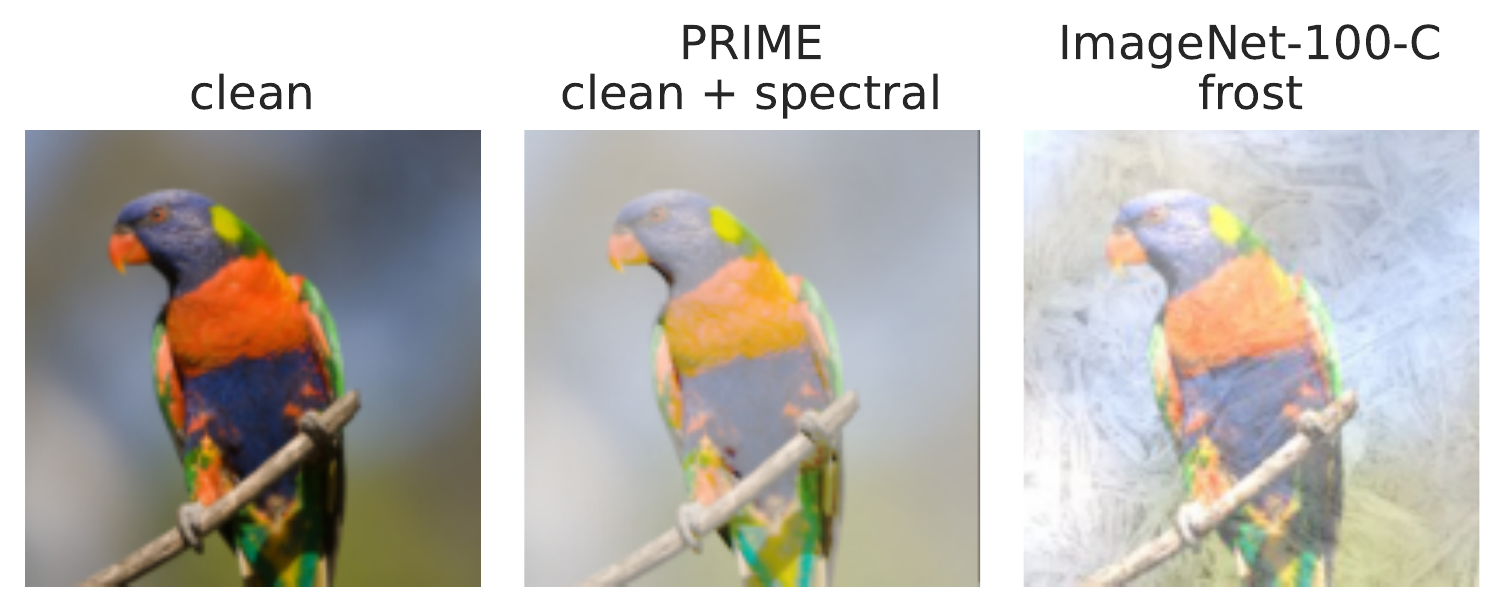}
        \captionsetup{justification=centering}
        \vspace*{-4mm}
        \caption{$\texttt{clean+}\texttt{spectral}$ \\$\approx\texttt{frost}$}
        \label{fig:add-mixing-examples:a1} 
    \end{subfigure}
    \begin{subfigure}[t]{0.395\columnwidth}
        \centering
        \footnotesize
        \includegraphics[width=\linewidth]{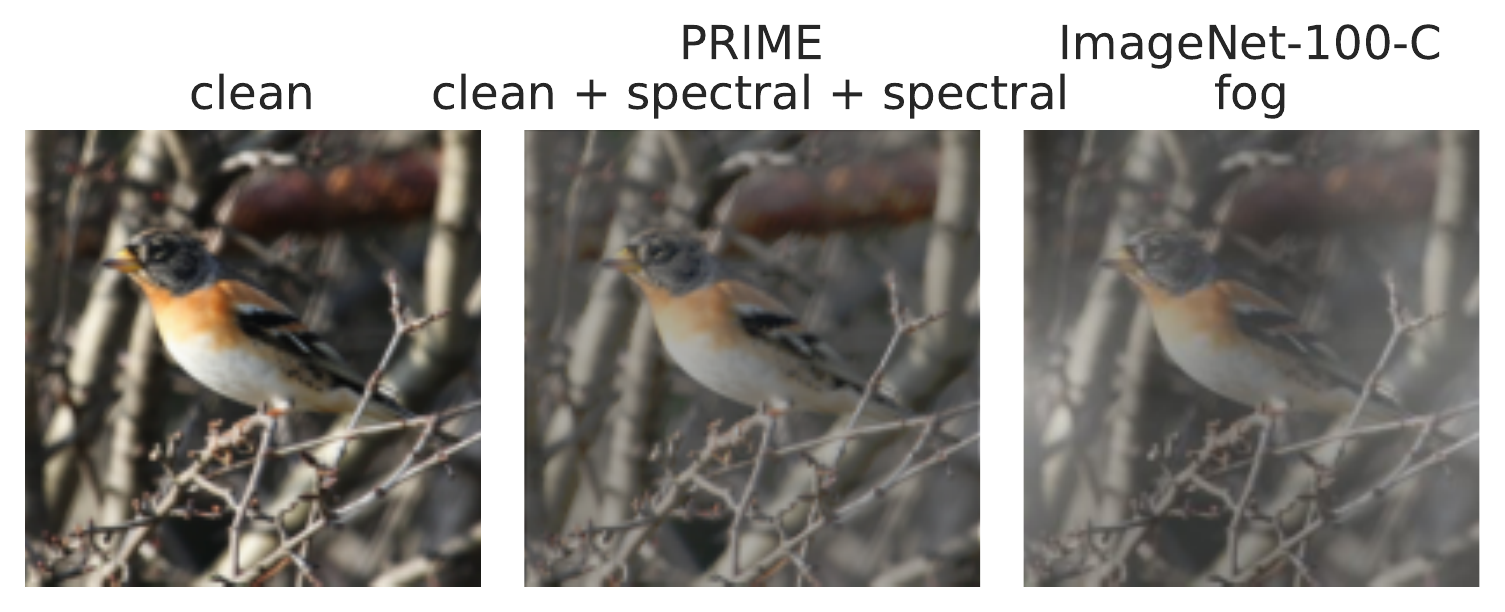}
        \captionsetup{justification=centering}
        \vspace*{-4mm}
        \caption{$\texttt{clean+}\texttt{spectral}$ \\$\texttt{+spectral}\approx\texttt{fog}$}
        \label{fig:add-mixing-examples:b1} 
    \end{subfigure}
    \begin{subfigure}[t]{0.395\columnwidth}
        \centering
        \footnotesize
        \includegraphics[width=\linewidth]{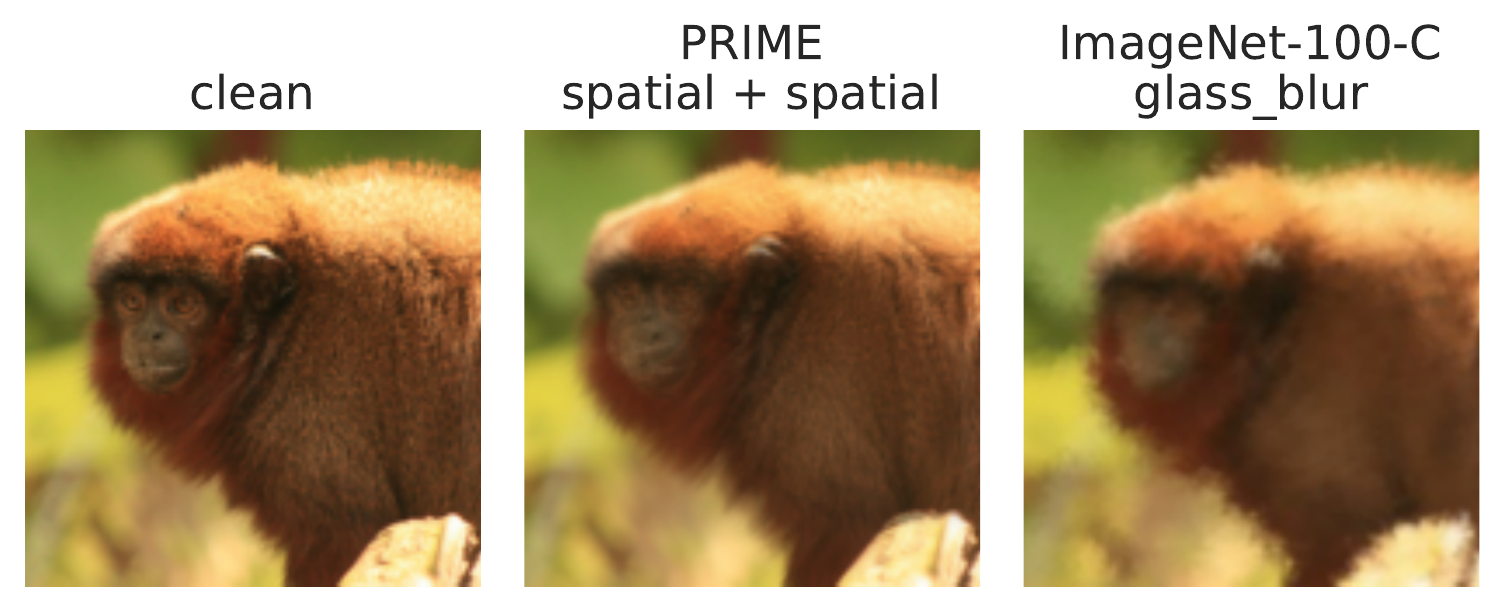}
        \captionsetup{justification=centering}
        \vspace*{-4mm}
        \caption{$\texttt{spatial+}\texttt{spatial}$ \\$\approx\texttt{glass\_blur}$}
        \label{fig:add-mixing-examples:c1} 
    \end{subfigure}
    \begin{subfigure}[t]{0.395\columnwidth}
        \centering
        \footnotesize
        \includegraphics[width=\linewidth]{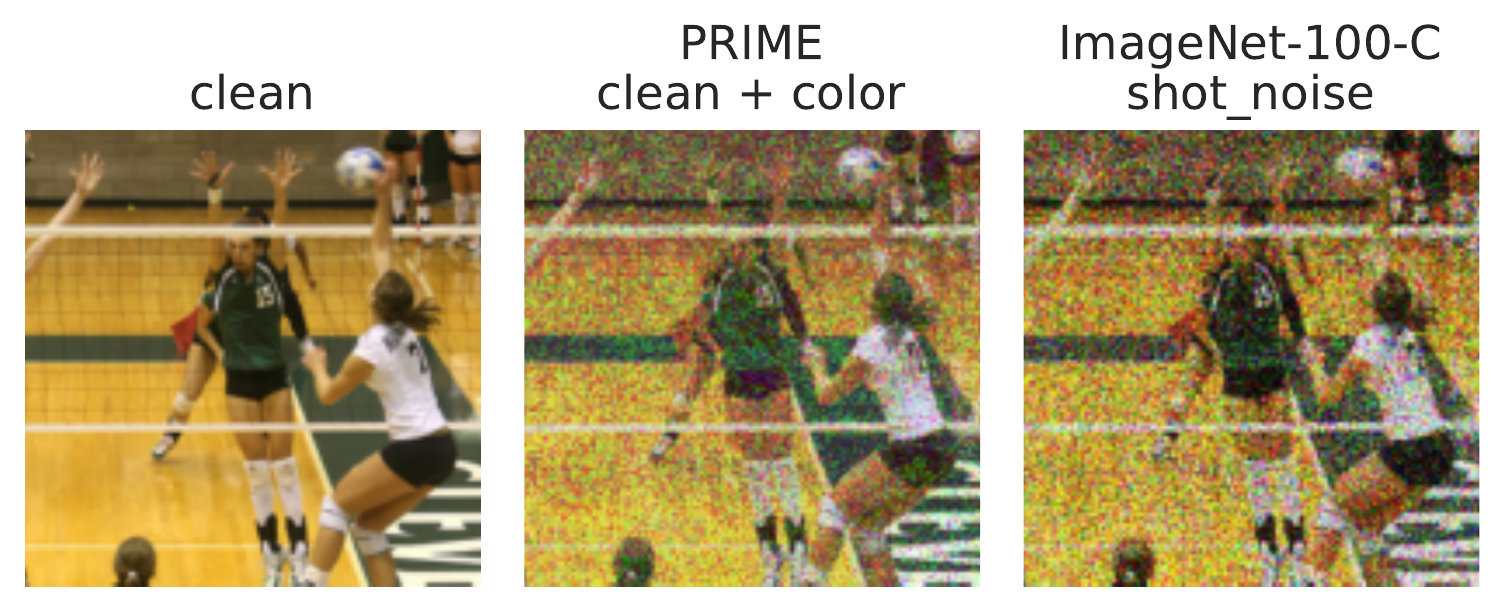}
        \captionsetup{justification=centering}
        \vspace*{-4mm}
        \caption{$\texttt{clean+}\texttt{color}$ \\$\approx\texttt{shot\_noise}$}
        \label{fig:add-mixing-examples:d1} 
    \end{subfigure}
    \begin{subfigure}[t]{0.395\columnwidth}
        \centering
        \footnotesize
        \includegraphics[width=\linewidth]{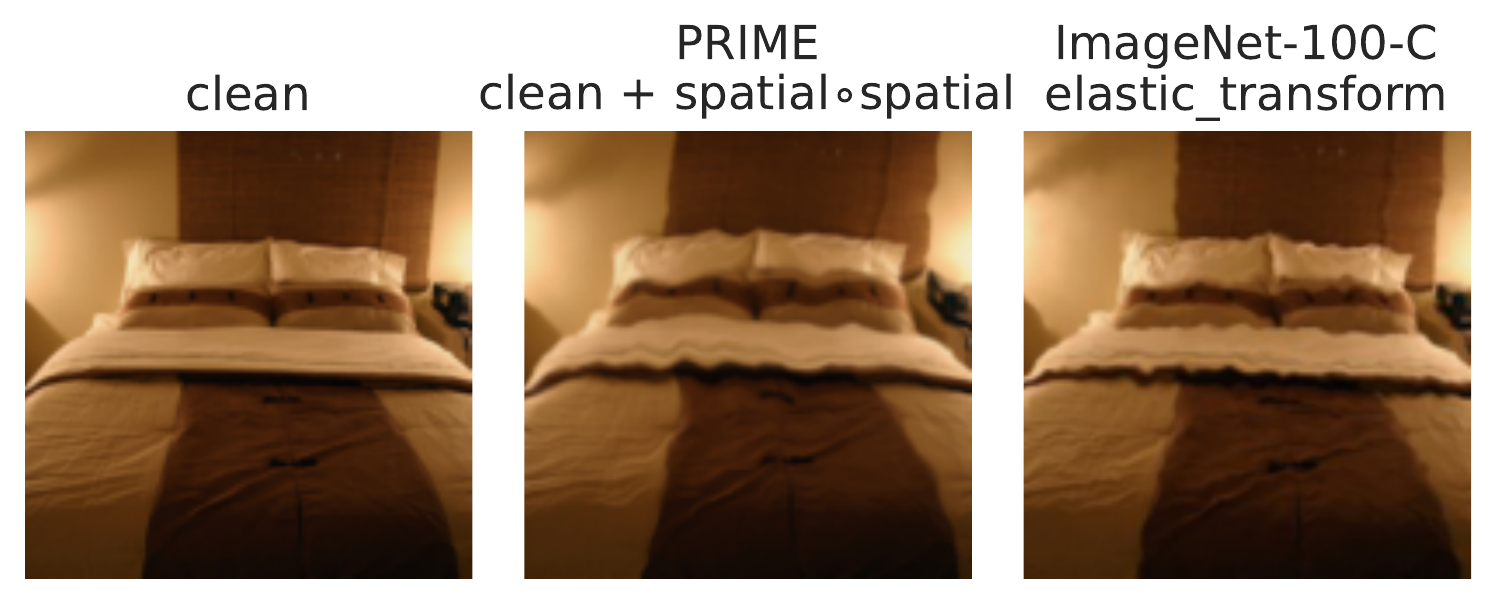}
        \captionsetup{justification=centering}
        \vspace*{-4mm}
        \caption{$\texttt{clean+spatial}\circ\texttt{spatial}$ \\$\approx\texttt{elastic\_transform}$}
        \label{fig:add-mixing-examples:e1} 
    \end{subfigure}
    \begin{subfigure}[t]{0.395\columnwidth}
        \centering
        \footnotesize
        \includegraphics[width=\linewidth]{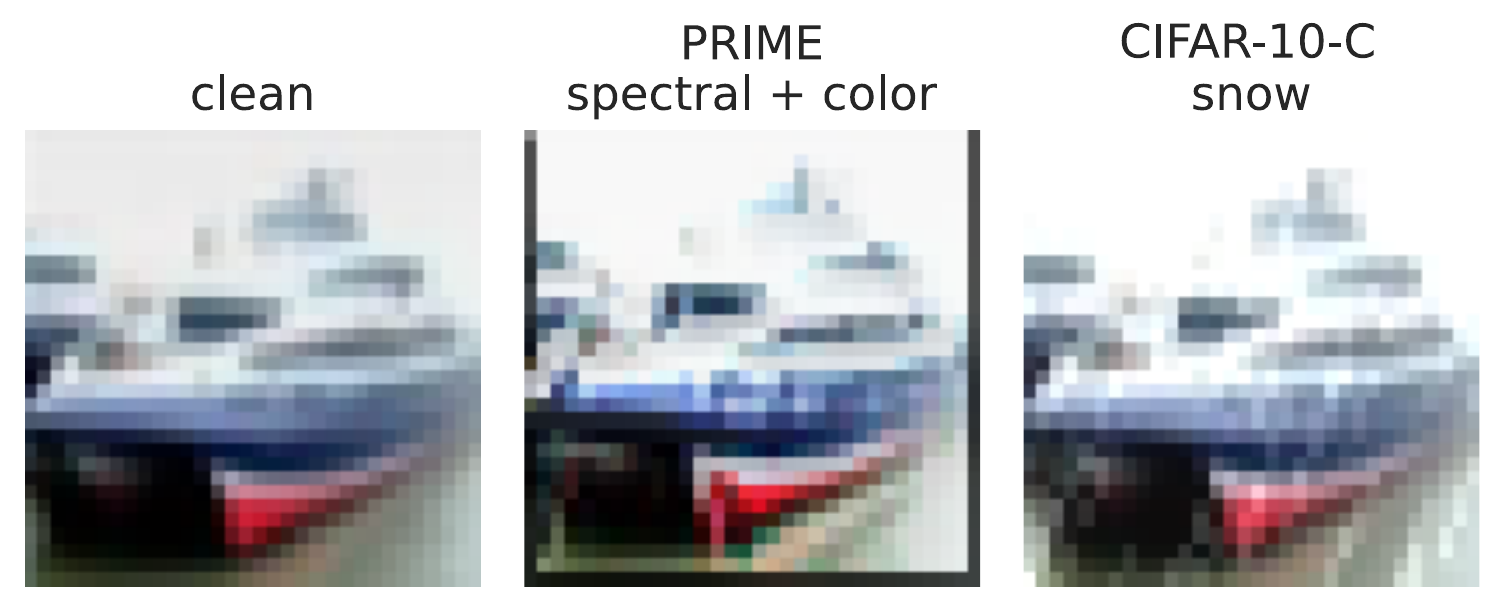}
        \captionsetup{justification=centering}
        \vspace*{-4mm}
        \caption{$\texttt{spectral+}\texttt{color}$ \\$\approx\texttt{snow}$}
        \label{fig:add-mixing-examples:f1} 
    \end{subfigure}
    
    \vspace{3pt}\rule{0.95\columnwidth}{0.4pt}
    \vspace{3pt}
    AugMix
    
    \begin{subfigure}[t]{0.395\columnwidth}
        \centering
        \footnotesize
        \includegraphics[width=\linewidth]{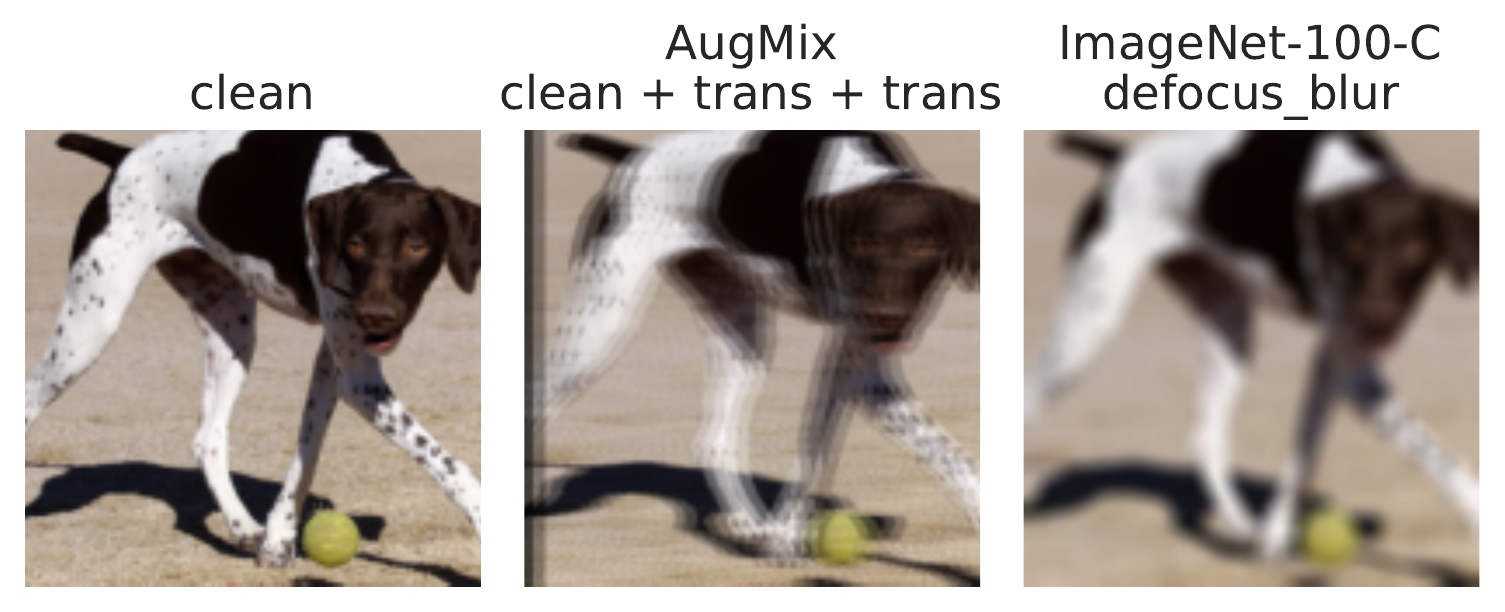}
        \captionsetup{justification=centering}
        \vspace*{-4mm}
        \caption{$\texttt{clean+}\texttt{translate}$ \\$\texttt{+translate}\approx\texttt{defocus\_blur}$}
        \label{fig:add-mixing-examples:a2} 
    \end{subfigure}
    \begin{subfigure}[t]{0.395\columnwidth}
        \centering
        \footnotesize
        \includegraphics[width=\linewidth]{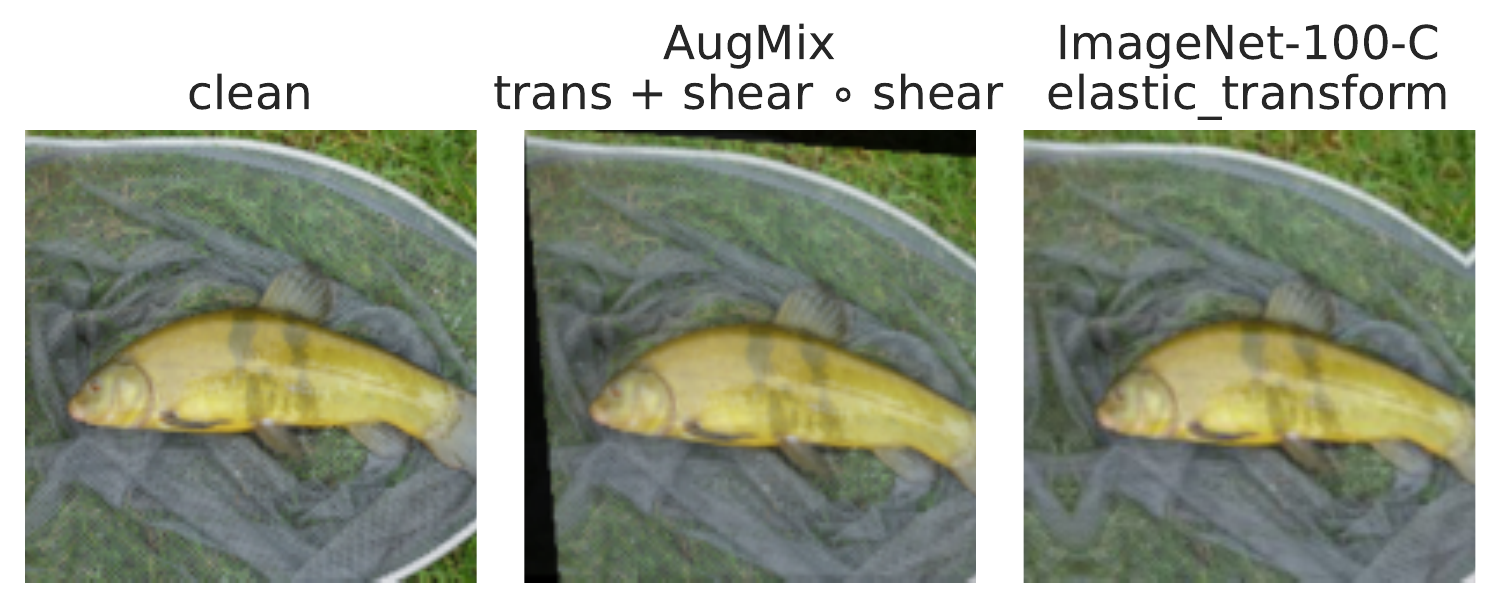}
        \captionsetup{justification=centering}
        \vspace*{-4mm}
        \caption{$\texttt{translate}\texttt{+shear}\circ\texttt{shear}$ \\$\approx\texttt{elastic\_transform}$}
        \label{fig:add-mixing-examples:b2} 
    \end{subfigure}
    \begin{subfigure}[t]{0.395\columnwidth}
        \centering
        \footnotesize
        \includegraphics[width=\linewidth]{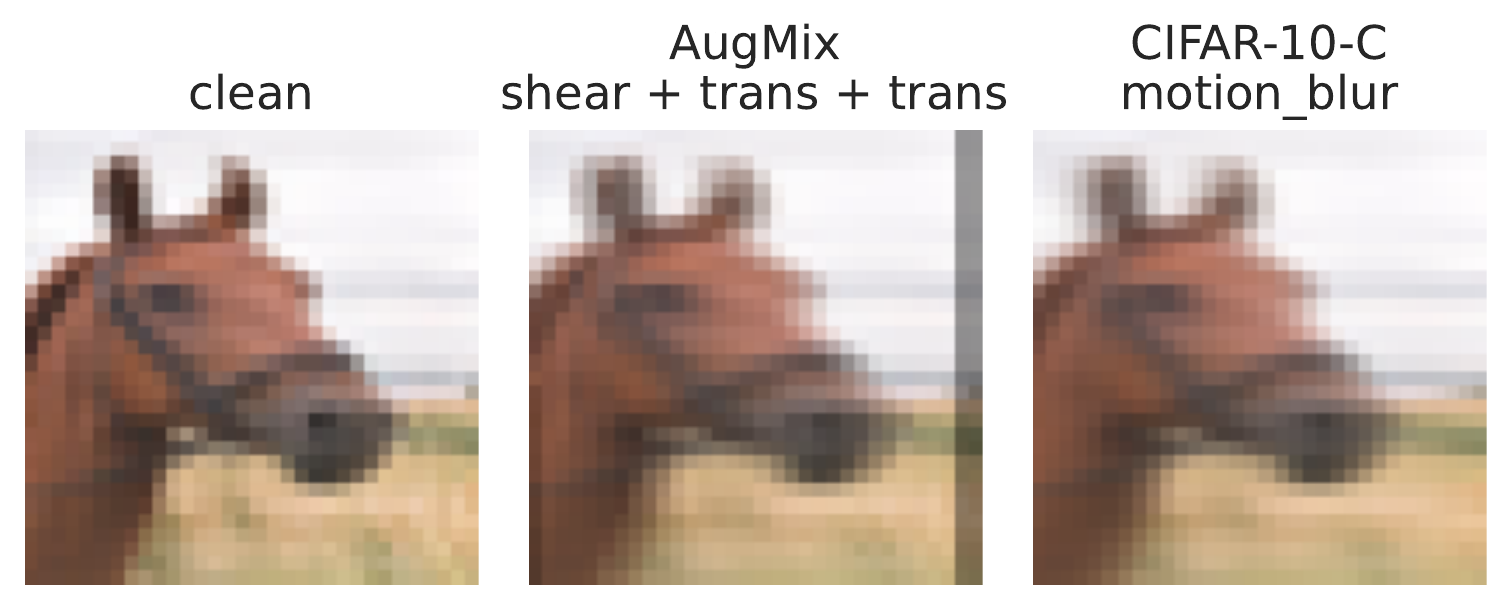}
        \captionsetup{justification=centering}
        \vspace*{-4mm}
        \caption{$\texttt{shear+}\texttt{translate+}$\\
        $\texttt{translate} \approx \texttt{motion\_blur}$}
        \label{fig:add-mixing-examples:c2}
    \end{subfigure}
    \begin{subfigure}[t]{0.395\columnwidth}
        \centering
        \footnotesize
        \includegraphics[width=\linewidth]{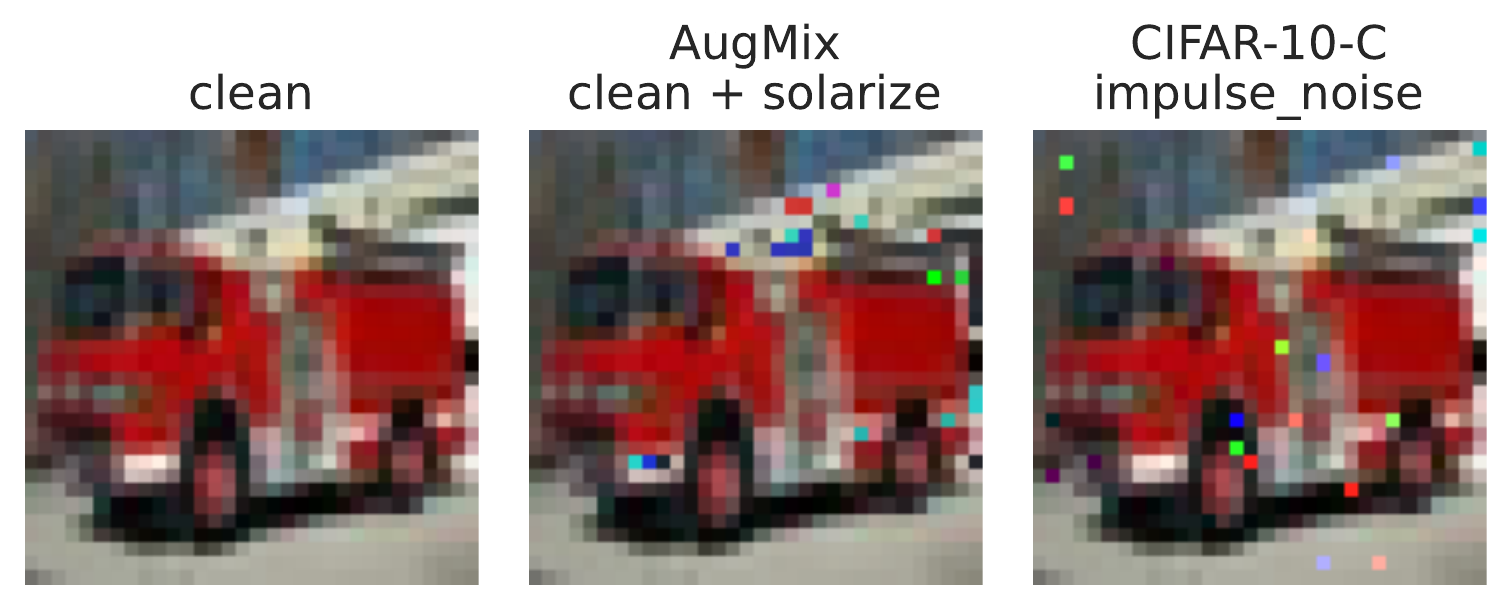}
        \captionsetup{justification=centering}
        \vspace*{-4mm}
        \caption{$\texttt{clean+}\texttt{solarize}$ \\$\approx\texttt{impulse\_noise}$}
        \label{fig:add-mixing-examples:d2} 
    \end{subfigure}
    \caption{The mixing procedure creates distorted images that look visually similar to the test-time corruptions. In each example (CIFAR-10/ImageNet-100), we show the clean image, the PRIME/AugMix augmented image and the corresponding common corruption that resembles the image produced by mixing. We also report the mixing combination used for recreating the corruption. $\circ$ stands for composition and $\texttt{+}$ represents convex combination (mixing). (Top 3 rows): PRIME, and (Last 2 rows): AugMix.}
    \vspace{-0.8em}
\label{fig:add-mixing-examples}
\end{figure}

Continuing \cref{subsec:role_of_mixing}, we present additional examples in \cref{fig:add-mixing-examples} to demonstrate the significance of mixing in PRIME. We observe that the mixing procedure is capable of constructing augmented images that look perceptually similar to common corruptions. To illustrate this, we provide several examples in \cref{fig:add-mixing-examples} for PRIME (upper half) and AugMix (lower half) on CIFAR-10 and ImageNet-100. As shown in \cref{fig:add-mixing-examples:a1,,fig:add-mixing-examples:b1}, mixing spectral transformations with the clean images tends to create weather-like artefacts resembling frost and fog respectively. Carefully combining clean and spatially transformed images produces blurs (\cref{fig:add-mixing-examples:c1}) and even elastic transform (\cref{fig:add-mixing-examples:e1}). Moreover, blending color augmentation with clean image produces shot noise as evident in \cref{fig:add-mixing-examples:d1}; Whereas spectral\texttt{+}color transformed image looks similar to snow corruption (\cref{fig:add-mixing-examples:f1}). All these observations explain the good performance of PRIME on the respective corruptions. 

Apart from the mixing in PRIME, the mixing in AugMix also plays a crucial role in its performance. In fact, a combination of translate and shear operations with the clean image create blur-like modifications that resemble defocus blur (\cref{fig:add-mixing-examples:a2}) and motion blur (\cref{fig:add-mixing-examples:c2}). This answers why AugMix excels at blur corruptions and is even better than DeepAugment against blurs (cf. \cref{tab:results-per-corruption-imagenet}). In addition, on CIFAR-10, notice that mixing solarize and clean produces impulse noise-like modifications (\cref{fig:add-mixing-examples:d2}), which justifies the improvements on noise attained by AugMix (refer \cref{tab:results-per-corruption-cifar}).

\clearpage
\newpage

\section{SimCLR nearest neighbours}
\label{app:simclr_NNs}

\begin{figure}[!ht]
\centering
\includegraphics[width=0.7\columnwidth]{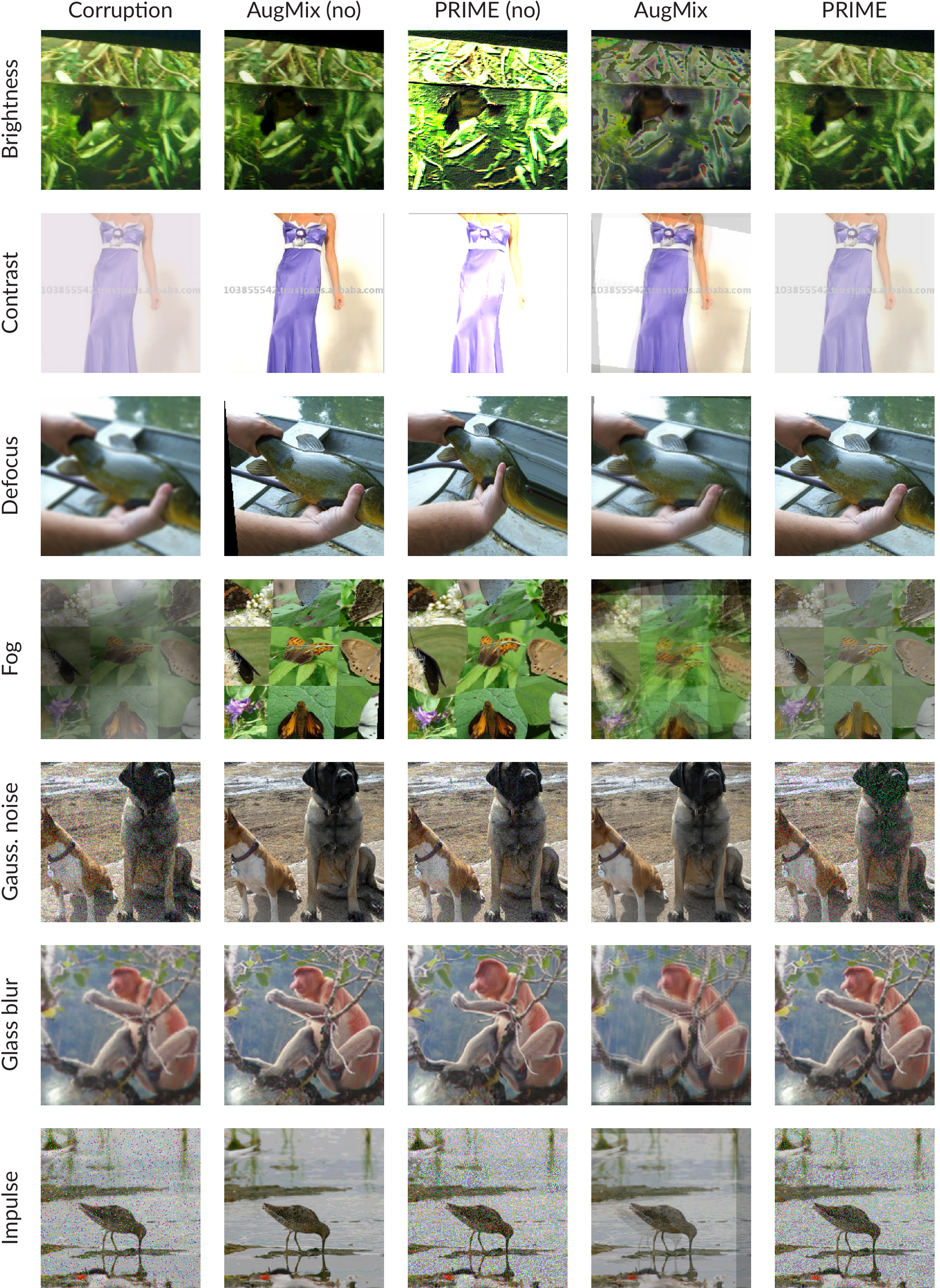}
\caption{Examples of nearest neighbours in SimCLRv2 embedding space. Columns: (first): the common corruption; (second): AugMix transformations (no mixing); (third): PRIME transformations (no mixing); (fourth): AugMix; (fifth): PRIME.}
\vspace{-0.8em}
\label{fig:simclr_NNs}
\end{figure}

Regarding the minimum distances in the SimCLRv2 embedding space of \cref{tab:simclr_distances}, we also provide in \cref{fig:simclr_NNs} some visual examples of the nearest neighbours of each method. In general, we observe that indeed smaller distance in the embedding space typically corresponds to closer visual similarity in the input space, with PRIME generating images that resemble more the corresponding common corruptions, compared to AugMix. Nevertheless, we also notice that for ``Blurs'' AugMix generates images that are more visually similar to the corruptions than PRIME, an observation that is on par with the lower performance of PRIME (without JSD) on blur corruptions (cf. \cref{tab:results-per-corruption-imagenet}) compared to AugMix.

\section{Cosine distance statistics}
\label{app:simclr_cosine_stats}

Recall that in \cref{tab:simclr_distances} we provide the average and the median of the minimum cosine distances computed in the SimCLRv2 embedding space. We now provide in \cref{tab:simclr_percentiles} the values for different percentiles of these distances. We observe that the behaviour is consistent across different percentiles: PRIME (with or without mixing) is always producing feature representations that are more similar to the common corruptions, compared to any version of AugMix. Note also that for smaller percentiles ($5\%,10\%,25\%$) it seems that PRIME without mixing reaches even lower values than PRIME. However, the difference with respect to PRIME can be considered as insignificant since it is in the order of $10^{-5}$ (note that all values in the table are in the order of $10^{-3}$); while a larger population of images ($>1000$) would potentially smooth out this difference.

\begin{table}[tb]
    \centering
    \footnotesize
    \caption{Percentiles of the minimum cosine distances in the ResNet-50 SimCLRv2 embedding space between $100$ augmented samples from $1000$ ImageNet images, and their corresponding common corruptions.}
    \aboverulesep=0ex
    \belowrulesep=0ex
    \begin{tabular}{lccccc}
        \MyToprule{1-6}
        \rule{0pt}{1.1EM}
        \multirow{2}{*}{Method} & \multicolumn{5}{c}{Min. cosine distance \footnotesize{($\times10^{-3}$)} ($\downarrow$)} \\
        \MyMidrule{2-6}
        & $5\%$ & $10\%$ & $25\%$ & $50\%$ & $75\%$ \\
        \midrule
        None (clean) & 0.33 & 0.64 & 1.97 & 6.43 & 17.44 \\
        \midrule
        AugMix (w/o mix) & 0.17 & 0.31 & 1.04 & 3.55 & 10.71 \\
        PRIME (w/o mix) & 0.04 & 0.07 & 0.24 & 1.87 & \multicolumn{1}{c}{7.11} \\
        \midrule
        AugMix & 0.11 & 0.21 & 0.69 & 2.61 & \multicolumn{1}{c}{8.37}\\
        PRIME & 0.08 & 0.12 & 0.32 & 1.61 & \multicolumn{1}{c}{5.76} \\
        \bottomrule
    \end{tabular}
\vspace{-0.8em}
\label{tab:simclr_percentiles}
\end{table}

\section{Embedding space visualization}
\label{app:embedding-visual}

\begin{figure}[!ht]
    \begin{center}
    \includegraphics[width=0.8\linewidth]{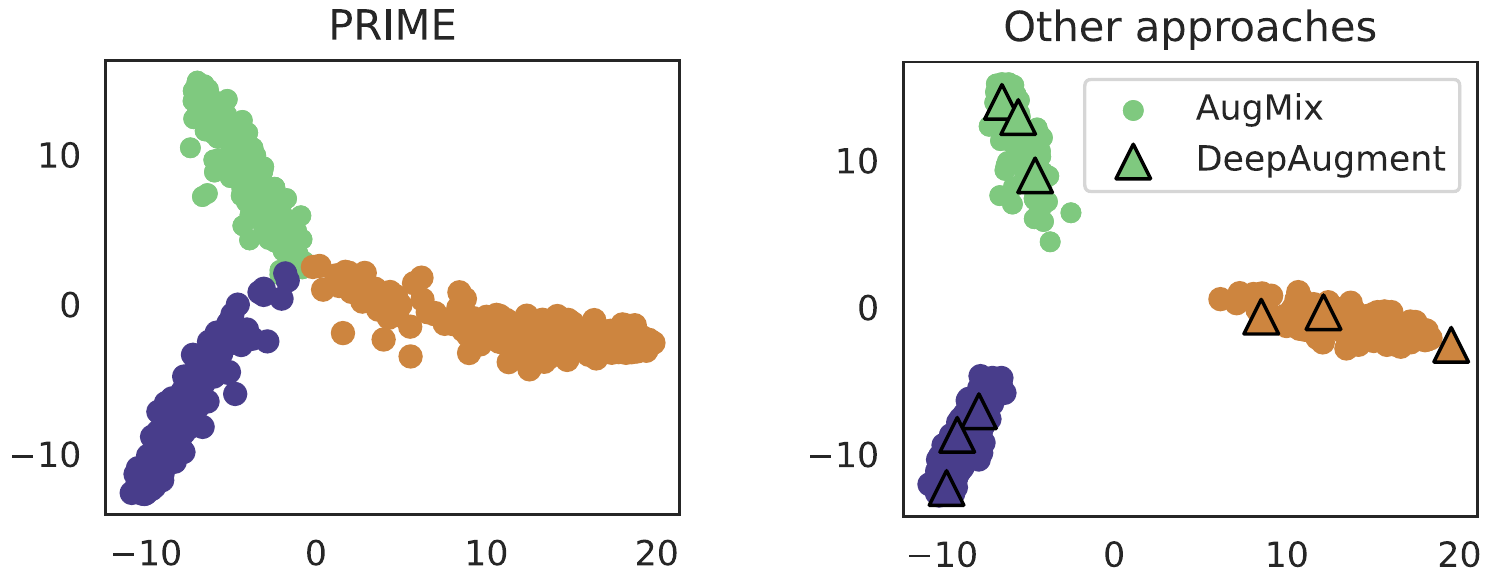}
    \end{center}
    \caption{Projections of augmentations generated by different methods on the embedding space of a ResNet-50.}
    \label{fig:prime-diversity-pca-feat-space}
\end{figure}

To qualitatively compare how diverse are the augmentations of PRIME with respect to other methods, we can follow the procedure in~\cite{augmax2021}. We randomly select 3 images from ImageNet, each one belonging to a different class. For each image, we generate 100 transformed instances using AugMix and PRIME, while with DeepAugment we can only use the original images and the 2 transformed instances that are pre-generated with the EDSR and CAE image-to-image networks that DeepAugment uses. Then, we pass the transformed instances of each method through a ResNet-50 pre-trained on ImageNet and extract the features of its embedding space. On the features extracted for each method, we perform PCA and then visualize the projection of the features onto the first two principal components. We visualize the projected augmented space in \cref{fig:prime-diversity-pca-feat-space}, which demonstrates that PRIME generates more diverse (larger variance) features than AugMix and DeepAugment.

\clearpage
\newpage

\section{Performance per corruption}
\label{app:performance-per-corruption}

% Tables from next section
\begin{table*}[bp]
\centering
\footnotesize
\caption{Per-corruption accuracy of different methods on C-10/100 (ResNet-18).}
\resizebox{\linewidth}{!}{%
    \centering
    \aboverulesep=0ex
    \belowrulesep=0ex
    % \footnotesize
    \begin{tabular}{clcc|ccc|cccc|cccc|cccc}
        \toprule
        \multirow{2}{*}{Dataset} & \multicolumn{1}{c}{\multirow{2}{*}{Method}} & \multicolumn{1}{c}{\multirow{2}{*}{Clean}} & \multicolumn{1}{c|}{\multirow{2}{*}{CC}} & \multicolumn{3}{c|}{Noise} & \multicolumn{4}{c|}{Blur} & \multicolumn{4}{c|}{Weather} & \multicolumn{4}{c}{Digital} \\
        % \MyMidrule{4-6}
        % \rule{0pt}{1.1EM}
        & & & & Gauss. & Shot & Impulse & Defoc. & Glass & Motion & Zoom & Snow & Frost & Fog & Bright. & Contr. & Elastic & Pixel. & JPEG\\
        \midrule
        \multirow{3}{*}{C-10} & Standard & 95.0 & 74.0 & 45.1 & 58.7 & 54.9 & 83.2 & 53.3 & 76.9 & 79.1 & 83.1 & 79.3 & 89.0 & 93.6 & 76.3 & 83.9 & 75.1 & 77.9\\
        & AugMix & 95.2 & 88.6 & 79.3 & 84.8 & 85.8 & 94.1 & 78.9 & 92.4 & 93.4 & 89.7 & 89.0 & 91.9 & 94.3 & 90.5 & 90.5 & 87.6 & 87.5\\
        & PRIME & 94.2 & 89.8 & 86.9 & 88.1 & 88.6 & 92.6 & 85.3 & 90.8 & 92.2 & 89.3 & 90.5 & 89.8 & 93.7 & 92.4 & 90.1 & 88.1 & 88.8\\
        \midrule
        \multirow{3}{*}{C-100}& Standard & 76.7 & 51.9 & 25.3 & 33.7 & 26.6 & 60.8 & 47.1 & 55.5 & 57.6 & 60.8 & 56.2 & 62.5 & 72.2 & 53.2 & 63.4 & 50.1 & 52.7\\
        & AugMix & 78.2 & 64.9 & 46.7 & 55.1 & 60.6 & 76.2 & 47.3 & 72.6 & 74.3 & 67.4 & 64.4 & 69.9 & 75.5 & 67.4 & 69.6 & 64.9 & 61.8\\
        & PRIME & 78.4 & 68.2 & 59.0 & 62.1 & 68.1 & 74.0 & 58.3 & 70.5 & 72.3 & 68.9 & 68.5 & 69.8 & 76.8 & 74.4 & 70.1 & 65.5 & 64.4\\
        \bottomrule
    \end{tabular}}
    \label{tab:results-per-corruption-cifar}
\end{table*}

\begin{table*}[tbp]
\centering
\footnotesize
\caption{Per-corruption accuracy of different methods on IN-100 (ResNet-18) and IN (ResNet-50). $^\dagger$ indicates that JSD consistency loss is not used. $^*$Models taken from~\protect\cite{robustbench2021}.}
\resizebox{\linewidth}{!}{%
    \centering
    \aboverulesep=0ex
    \belowrulesep=0ex
    \begin{tabular}{clcc|ccc|cccc|cccc|cccc}
        \toprule
        \multirow{2}{*}{Dataset} & \multicolumn{1}{c}{\multirow{2}{*}{Method}} & \multicolumn{1}{c}{\multirow{2}{*}{Clean}} & \multicolumn{1}{c|}{\multirow{2}{*}{CC}} & \multicolumn{3}{c|}{Noise} & \multicolumn{4}{c|}{Blur} & \multicolumn{4}{c|}{Weather} & \multicolumn{4}{c}{Digital} \\
        % \MyMidrule{4-6}
        % \rule{0pt}{1.1EM}
        & & & & Gauss. & Shot & Impulse & Defoc. & Glass & Motion & Zoom & Snow & Frost & Fog & Bright. & Contr. & Elastic & Pixel. & JPEG\\
        \midrule
        \multirow{6}{*}{IN-100} & Standard & 88.0 & 49.7 & 30.9 & 29.0 & 22.0 & 45.6 & 44.6 & 50.4 & 53.9 & 43.8 & 46.2 & 50.5 & 78.6 & 42.9 & 68.8 & 68.0 & 70.6 \\
        & AugMix & 88.7 & 60.7 & 45.2 & 45.8 & 43.4 & 58.7 & 53.3 & 69.5 & 71.0 & 49.1 & 52.7 & 60.2 & 80.7 & 59.6 & 73.3 & 73.6 & 74.7\\
        & DA & 86.3 & 67.7 & 76.3 & 75.6 & 75.7 & 64.2 & 61.7 & 61.3 & 62.7 & 54.4 & 62.8 & 55.7 & 81.6 & 49.7 & 69.9 & 83.3 & 80.6\\
        & PRIME & 85.9 & 71.6 & 80.6 & 80.0 & 80.1 & 57.2 & 66.3 & 66.2 & 68.2 & 61.5 & 68.2 & 57.2 & 81.2 & 68.3 & 73.7 & 82.9 & 81.9\\
        \cmidrule{2-19}
        & DA\texttt{+}AugMix & 86.5 & 73.1 & 75.2 & 75.8 & 74.9 & 74.1 & 68.5 & 76.0 & 72.1 & 59.9 & 66.8 & 61.4 & 82.1 & 72.4 & 73.1 & 83.8 & 81.1\\
        & DA\texttt{+}PRIME & 84.9 & 74.9 & 81.1 & 80.9 & 81.2 & 70.5 & 74.2 & 72.0 & 71.5 & 66.3 & 73.6 & 56.6 & 81.9 & 72.8 & 74.8 & 83.4 & 82.3\\
        \midrule
        \multirow{7}{*}{IN} & Standard$^*$ & 76.1 & 39.2 & 29.3 & 27.0 & 23.8 & 38.8 & 26.8 & 38.7 & 36.2 & 32.5 & 38.1 & 45.4 & 68.0 & 39.0 & 45.3 & 44.8 & 53.4  \\
        & AugMix$^*$ & 77.5 & 48.3 & 40.6 & 41.1 & 37.7 & 47.7 & 34.9 & 53.5 & 49.0 & 39.9 & 43.8 & 47.1 & 69.5 & 51.1 & 52.0 & 57.0 & 60.3 \\
        & DA$^*$ & 76.7 & 52.6 & 56.6 & 54.9 & 56.3 & 51.7 & 40.1 & 48.7 & 39.5 & 44.2 & 50.3 & 52.1 & 71.1 & 48.3 & 50.9 & 65.5 & 59.3 \\
        & PRIME$^\dagger$ & 77.0 & 55.0 & 61.9 & 60.6 & 60.9 & 47.6 & 39.0 & 48.4 & 46.0 & 47.4 & 50.8 & 54.1 & 71.7 & 58.2 & 56.3 & 59.5 & 62.2\\
        % & (w/o JSD) & &\\
        \cmidrule{2-19}
        & DA\texttt{+}AugMix & 75.8 & 58.1 & 59.4 & 59.6 & 59.1 & 59.0 & 46.8 & 61.1 & 51.5 & 49.4 & 53.3 & 55.9 & 70.8 & 58.7 & 54.3 & 68.8 & 63.3 \\
        & DA\texttt{+}PRIME$^\dagger$ & 75.5 & 59.9 & 67.4 & 67.2 & 66.8 & 56.2 & 47.5 & 54.3 & 47.3 & 52.8 & 56.4 & 56.3 & 71.7 & 62.3 & 57.3 & 70.3 & 65.1 \\
        % & (w/o JSD) & &\\ 
        \bottomrule
    \end{tabular}}
    \label{tab:results-per-corruption-imagenet}
\end{table*}

Beyond the average corruption accuracy that we report in \cref{tab:results-main}, we also provide here the performance of each method on the individual corruptions. The results on CIFAR-10/100 and ImageNet/ImageNet-100 are shown on \cref{tab:results-per-corruption-cifar} and \cref{tab:results-per-corruption-imagenet} respectively. Compared to AugMix on CIFAR-10/100, the improvements from PRIME are mostly observed against Gaussian noise ($+7.6\%/12.3\%$), shot noise ($+3.3\%/7.0\%$), glass blur ($+6.4\%/11.0\%$) and JPEG compression ($+1.3\%/2.6\%$). These results show that PRIME can really push the performance against certain corruptions in CIFAR-10/100-C despite the fact that AugMix is already good on these datasets. However, AugMix turns out to be slightly better than PRIME against impulse noise, defocus blur and motion blur modifications; all of which have been shown to be resembled by AugMix created images (see \cref{fig:add-mixing-examples}). With ImageNet-100, PRIME enhances the diversity of augmented images, and leads to general improvements against all corruptions except certain blurs.
On ImageNet, we observe that, in comparison to DeepAugment, the supremacy of PRIME is reflected on almost every corruption type, except some blurs and pixelate corruptions where DeepAugment is slightly better. When PRIME is used in conjunction with DeepAugment, compared to AugMix combined with DeepAugment, our method seems to lack behind only on blurs, while on the rest of the corruptions achieves higher robustness.
% Regarding ImageNet, we observe that, compared to DeepAugment, the main added value of PRIME is reflected on zoom blurs ($+5.2\%$), contrast ($+6.5\%$) and elastic transforms ($+3.8\%$). However, when PRIME is combined with DeepAugment then, compared to AugMix combined with DeepAugment, our method seems to help on improving on all the noises by $+4.1\%$ on average.

\section{Performance per severity level}

We also want to investigate the robustness of each method on different severity levels of the corruptions. The results for CIFAR-10/100 and ImageNet/ImageNet-100 are presented in \cref{tab:results-per-severity-cifar} and \cref{tab:results-per-severity-imagenet} respectively. With CIFAR-10/100, PRIME predominantly helps against corruptions with maximal severity and yields $+3.9\%$ and $+7.1\%$ gains on CIFAR-10 and CIFAR-100 respectively. Besides on ImageNet-100, PRIME again excels at corruptions with moderate to higher severity. This observations also holds when PRIME is employed in concert with DeepAugment. With ImageNet too this trend continues, and we observe that, compared to DeepAugment, PRIME improves significantly on corruptions of larger severity ($+3.4\%$ and $+5.5\%$ on severity levels 4 and 5 respectively). Also, this behaviour is consistent even when PRIME is combined with DeepAugment and is compared to DeepAugment\texttt{+}AugMix, where we see that again on levels 4 and 5 there is a significant improvement of $+2.1\%$ and $+3.7\%$ respectively.
% Regarding ImageNet too this trend continues, and we observe that, compared to DeepAugment, PRIME improves mostly on corruptions of larger severity ($+1.9\%$). Also, this behaviour is consistent even when PRIME is combined with DeepAugment and is compared to AugMix combined with DeepAugment, where we see that again on the strongest severity level there is an improvement of $+1.6\%$.

\begin{table*}[!ht]
\centering
\footnotesize
\caption{Average accuracy for each corruption severity level of different methods on C-10 and C-100 (ResNet-18).}
\aboverulesep=0ex
\belowrulesep=0ex
\begin{tabular}{clcc|ccccc}
    \toprule
    \multirow{2}{*}{Dataset} & \multicolumn{1}{c}{\multirow{2}{*}{Method}} & \multicolumn{1}{c}{\multirow{2}{*}{Clean}} & \multicolumn{1}{c|}{\multirow{2}{*}{CC Avg.}} & \multicolumn{5}{c}{Severity} \\
    % \MyMidrule{4-6}
    % \rule{0pt}{1.1EM}
    & & & & 1 & 2 & 3 & 4 & 5\\
    \midrule
    \multirow{3}{*}{C-10} & Standard & 95.0 & 74.0 & 87.4 & 81.7 & 75.7 & 68.3 & 56.7\\
    & AugMix & 95.2 & 88.6 & 93.1 & 91.8 & 89.9 & 86.7 & 81.7\\
    & PRIME & 94.2 & 89.8 & 92.8 & 91.6 & 90.4 & 88.6 & 85.6\\
    \midrule
    \multirow{3}{*}{C-100}& Standard & 76.7 & 51.9 & 66.7 & 59.4 & 52.8 & 45.0 & 35.4\\
    & AugMix & 78.2 & 64.9 & 73.3 & 70.0 & 66.6 & 61.3 & 53.4\\
    & PRIME & 78.4 & 68.2 & 74.0 & 71.6 & 69.2 & 65.6 & 60.5\\
    \bottomrule
\end{tabular}
\label{tab:results-per-severity-cifar}
\end{table*}

\begin{table*}[t]
\centering
\footnotesize
\caption{Average accuracy for each corruption severity level of different methods on IN-100 (ResNet-18) and IN (ResNet-50). $^\dagger$ indicates that JSD consistency loss is not used. $^*$Models taken from \texttt{RobustBench}~\protect\cite{robustbench2021}.}
\aboverulesep=0ex
\belowrulesep=0ex
\begin{tabular}{clcc|ccccc}
    \toprule
    \multirow{2}{*}{Dataset} & \multicolumn{1}{c}{\multirow{2}{*}{Method}} & \multicolumn{1}{c}{\multirow{2}{*}{Clean}} & \multicolumn{1}{c|}{\multirow{2}{*}{CC Avg.}} & \multicolumn{5}{c}{Severity} \\
    % \MyMidrule{4-6}
    % \rule{0pt}{1.1EM}
    & & & & 1 & 2 & 3 & 4 & 5\\
    \midrule
    \multirow{6}{*}{IN-100} & Standard & 88.0 & 49.7 & 73.5 & 61.0 & 49.8 & 37.2 & 27.0\\
    & AugMix & 88.7 & 60.7 & 80.4 & 71.8 & 63.8 & 50.3 & 37.2\\
    & DA & 86.3 & 67.7 & 81.2 & 75.4 & 69.9 & 61.2 & 50.8\\
    & PRIME & 85.9 & 71.6 & 81.7 & 77.5 & 73.4 & 66.9 & 58.4\\
    \cmidrule{2-9}
    & DA\texttt{+}AugMix & 86.5 & 73.1 & 82.7 & 78.0 & 75.5 & 69.6 & 59.9\\
    & DA\texttt{+}PRIME & 84.9 & 74.9 & 82.0 & 78.7 & 76.4 & 71.8 & 65.5\\
    \midrule
    \multirow{7}{*}{IN} & Standard$^*$ & 76.1 & 39.2 & 60.6 & 49.8 & 39.8 & 27.7 & 18.0 \\
    & AugMix$^*$ & 77.5 & 48.3 & 66.7 & 58.3 & 51.1 & 39.1 & 26.5 \\
    & DA$^*$ & 76.7 & 52.6 & 69.0 & 61.7 & 55.4 & 44.9 & 32.1 \\
    & PRIME$^\dagger$ & 77.0 & 55.0 & 68.9 & 63.1 & 56.9 & 48.3 & 37.6 \\
    % & (w/o JSD) & &\\
    \cmidrule{2-9}
    & DA\texttt{+}AugMix & 75.8 & 58.1 & 70.3 & 64.5 & 60.5 & 53.0 & 42.2 \\
    & DA\texttt{+}PRIME$^\dagger$ & 75.5 & 59.9 & 70.8 & 66.3 & 61.6 & 55.1 & 45.9 \\
    % & (w/o JSD) & &\\ 
    \bottomrule
\end{tabular}
\label{tab:results-per-severity-imagenet}
\end{table*}

\section{Performance on other corruptions}
\label{app:other_datasets}

Finally, to examine the universality of PRIME, we evaluate the performance of our ImageNet-100 trained models against two other corrupted datasets: (i) ImageNet-100-$\overline{\text{C}}$ (IN-100-$\overline{\text{C}}$)~\cite{cbar2021}, and (ii) stylized ImageNet-100 (SIN-100)~\cite{styleimagenet2018}. While IN-100-$\overline{\text{C}}$ is composed of corruptions that are perceptually dissimilar to those in IN-100-C, stylized IN-100 only retains global shape information and discard local texture cues from IN-100 test images, via style transfer. Thus, it would be interesting test the performance of PRIME against these datasets since it would serve as a indicator for general corruption robustness of PRIME. More information about the corruption types contained in IN-100-$\overline{\text{C}}$ is available in the original paper~\cite{cbar2021}.

\begin{table*}[b!]
\centering
\footnotesize
\caption{Classification accuracy of different methods on IN-100-C, IN-100-$\overline{\text{C}}$ and Stylized IN-100 (SIN-100) with ResNet-18.}
\resizebox{\linewidth}{!}{
    \centering
    \aboverulesep=0ex
    \belowrulesep=0ex
    \footnotesize
    \begin{tabular}{lcc|c|cccccccccc|c}
        \toprule
        \addlinespace[0.1em]
        \multirow{2}{*}{Method} & \multirow{2}{*}{Clean} & IN-100-C & IN-100-$\overline{\text{C}}$ & \multicolumn{10}{c}{IN-100-$\overline{\text{C}}$} & \multicolumn{1}{|c}{\multirow{2}{*}{SIN-100}}\\
        & & Avg. & Avg. & BSmpl & Brown & Caustic & Ckbd & CSine & ISpark & Perlin & Plasma & SFreq & Spark &\\
        \midrule
        Standard & 88.0 & 49.7 & 55.1 & 47.6 & 71.3 & 70.1 & 66.4 & 29.5 & 45.7 & 72.1 & 34.6 & 34.9 & 78.4 & 18.8\\
        AugMix & 88.7 & 60.7 & 61.0 & 63.0 & 73.2 & 75.3 & 69.4 & 39.9 & 44.9 & 77.4 & 42.8 & 44.7 & 79.8 & 28.0\\
        DA & 86.3 & 67.7 & 63.8 & 77.1 & 76.6 & 72.6 & 60.9 & 42.9 & 44.3 & 78.0 & 43.4 & 64.5 & 77.8 & 29.9\\
        \midrule
        PRIME & 85.9 & 71.6 & 65.0 & 74.9 & 74.3 & 73.2 & 59.2 & 53.4 & 47.5 & 76.8 & 48.6 & 66.9 & 75.5 & 33.1\\
        \texttt{+}1.5x epochs & 86.1 & 72.5 & 65.9 & 77.1 & 75.6 & 74.1 & 59.4 & 54.0 & 46.3 & 77.6 & 50.4 & 67.7 & 76.4 & 34.1\\
        \bottomrule
    \end{tabular}}
\label{tab:results-per-corruption-cbar100}
\end{table*}

\cref{tab:results-per-corruption-cbar100} enumerates the classification accuracy of different standalone approaches against IN-100-$\overline{\text{C}}$ on average, individual corruptions in IN-100-$\overline{\text{C}}$ and SIN-100. We can see that PRIME surpasses AugMix and DeepAugment by $4\%$ and $1.2\%$ respectively on IN-100-$\overline{\text{C}}$. PRIME particularly helps against certain distortions such as blue noise sample (BSmpl), inverse sparkles and plasma noise. PRIME also works well against style-transferred images in SIN-100 and improves accuracy by $5.1\%$ over AugMix and $3.2\%$ over DeepAugment. Besides, the diversity of our method means that we can actually get a better performance by increasing the number of training epochs. With 1.5x training epochs, we observe about $1\%$ accuracy refinement on each benchmark. 

We also perform a similar analysis with ImageNet trained models and evaluate their robustness on three other distribution shift benchmarks: (i) IN-$\overline{\text{C}}$~\cite{cbar2021}, (ii) SIN~\cite{styleimagenet2018} as described previously and (iii) ImageNet-R (IN-R)~\cite{deepaugment2021}. ImageNet-R contains naturally occurring artistic renditions (e.g.,
paintings, embroidery, etc.) of objects from the ImageNet dataset. The classification accuracy achieved by different methods on these datasets is listed in \cref{tab:results-per-corruption-cbar}. On IN-$\overline{\text{C}}$, PRIME outperforms AugMix and DeepAugment by $3.1\%$ and $1.3\%$ respectively. Besides, PRIME also obtains competitive results on IN-R and SIN datasets. \balance Altogether, our empirical results indicate that the performance gains obtained by PRIME indeed translate to other corrupted datasets.

\begin{table*}[t!]
\centering
\footnotesize
\caption{Classification accuracy of different methods on IN-C, IN-$\overline{\text{C}}$, ImageNet-R (IN-R) and Stylized IN (SIN) with ResNet-50. $^\dagger$~indicates that JSD consistency loss is not used. $^*$Models taken from \texttt{RobustBench}~\protect\cite{robustbench2021}.}
\resizebox{\linewidth}{!}{%
    \centering
    \aboverulesep=0ex
    \belowrulesep=0ex
    \footnotesize
    \begin{tabular}{lcc|c|cccccccccc|cc}
        \toprule
        \addlinespace[0.1em]
        \multirow{2}{*}{Method} & \multirow{2}{*}{Clean} & IN-C & IN-$\overline{\text{C}}$ & \multicolumn{10}{c}{IN-$\overline{\text{C}}$} & \multicolumn{1}{|c}{\multirow{2}{*}{IN-R}} & \multicolumn{1}{c}{\multirow{2}{*}{SIN}}\\
        & & Avg. & Avg. & BSmpl & Brown & Caustic & Ckbd & CSine & ISpark & Perlin & Plasma & SFreq & Spark & &\\
        \midrule
        Standard$^*$ & 76.1 & 39.2 & 40.0 & 36.2 & 57.8 & 54.1 & 46.1 & 14.4 & 20.9 & 61.6 & 24.3 & 19.0 & 65.2 & 36.2 & 7.4\\
        AugMix$^*$ & 77.5 & 48.3 & 46.5 & 59.5 & 56.5 & 59.1 & 51.7 & 25.6 & 21.6 & 65.3 & 23.1 & 36.2 & 66.4 & 41.0 & 11.2\\
        DA$^*$ & 76.7 & 52.6 & 48.3 & 60.1 & 61.1 & 57.7 & 46.8 & 25.4 & 24.4 & 68.4 & 26.5 & 45.6 & 66.8 & 42.2 & 14.2\\
        PRIME$^\dagger$ & 77.0 & 55.0 & 49.6 & 59.5 & 61.4 & 60.1 & 48.1 & 26.9 & 28.3 & 66.5 & 36.4 & 41.9 & 66.5 & 42.2 & 14.0\\
        \bottomrule
    \end{tabular}}
\label{tab:results-per-corruption-cbar}
\end{table*}

\end{document}